\documentclass[sigconf,nonacm]{acmart}

\AtBeginDocument{%
  \providecommand\BibTeX{{%
    \normalfont B\kern-0.5em{\scshape i\kern-0.25em b}\kern-0.8em\TeX}}}

\usepackage{multirow}
\usepackage{bm}
\usepackage{multicol}
\usepackage{hyperref}

\newcommand{\anchor}{\bm{a}}
\newcommand{\positive}{\bm{p}}
\newcommand{\negative}{\bm{n}}

\def\eg{\emph{e.g.}}

\copyrightyear{2021}
\acmYear{2021}
\setcopyright{acmlicensed}\acmConference[MM '21]{Proceedings of the 29th ACM International Conference on Multimedia}{October 20--24, 2021}{Virtual Event, China}
\acmBooktitle{Proceedings of the 29th ACM International Conference on Multimedia (MM '21), October 20--24, 2021, Virtual Event, China}
\acmPrice{15.00}
\acmDOI{10.1145/3474085.3475465}
\acmISBN{978-1-4503-8651-7/21/10}

\acmSubmissionID{1666}

\begin{document}
\fancyhead{}

\title{Cross-Modal Retrieval and Synthesis (X-MRS): Closing the Modality Gap in Shared Representation Learning 
}

\author{Ricardo Guerrero}
\email{r.guerrero@samsung.com}
\affiliation{%
  \institution{Samsung AI Center}
  \city{Cambridge}
  \country{UK}
}

\author{Hai X. Pham}
\email{hai.xuanpham@samsung.com}
\affiliation{
  \institution{Samsung AI Center}
  \city{Cambridge}
  \country{UK}
}

\author{Vladimir Pavlovic}
\email{v.pavlovic@samsung.com}
\affiliation{%
  \institution{Samsung AI Center}
  \city{Cambridge}
  \country{UK}
}




\keywords{Cross-modal, retrieval, multilingual, synthesis}


\begin{abstract}
   Computational food analysis (CFA) naturally requires multi-modal evidence of a particular food, \eg, images, recipe text, etc. A key to making CFA possible is multi-modal shared representation learning, which aims to create a joint representation of the multiple views (text and image) of the data. In this work we propose a method for food domain cross-modal shared representation learning that preserves the vast semantic richness present in the food data. Our proposed method employs an effective transformer-based multilingual recipe encoder coupled with a traditional image embedding architecture. Here, we propose the use of imperfect multilingual translations to effectively regularize the model while at the same time adding functionality across multiple languages and alphabets. Experimental analysis on the public Recipe1M dataset shows that the representation learned via the proposed method significantly outperforms the current state-of-the-arts (SOTA) on retrieval tasks. Furthermore, the representational power of the learned representation is demonstrated through a generative food image synthesis model conditioned on recipe embeddings. Synthesized images can effectively reproduce the visual appearance of paired samples, indicating that the learned representation captures  the joint semantics of both the textual recipe and its visual content, thus narrowing the modality gap.\footnote{Code is available at \href{https://github.com/SamsungLabs/X-MRS}{https://github.com/SamsungLabs/X-MRS}}
\end{abstract}

\maketitle

\section{Introduction}
Computational food analysis (CFA) broadly refers to methods that attempt automating food understanding, and as such, it has recently received increased attention, in part due to its importance in health and general wellbeing. For instance, CFA  can play an important role in assessing and learning the functional similarity and interaction of ingredients, cooking methods and meal preferences, while aiding in computational meal preparation and planning~\cite{teng2012recipe,helmy2015health}. However, despite recent efforts CFA still poses specific and difficult challenges due to the highly heterogeneous and complex nature of the cooking transformation process. Further to this, a particular modality may offer only a partial ``view''of the item. For example, a cooking recipe often describes elements that can easily be occluded in the visual depiction of a cooked dish, and/or come in a variety of colors, forms and textures. Recent approaches that aim to bridge the gap between visual and textual representations of food items do so by learning a shared latent space \cite{salvador2017,chen2018,carvalho2018,wang2019,marin2019,Fu2020,Fain2019,Zhu2019,salvador2021,wang2020}. Here, paired multi-modal representations (also called embeddings) are derived from different sources (\eg, images and text). The objective of these approaches is to find a common modality independent grounding language, thus allowing ``translation'' between modalities. However, as learning effective decoders into high-dimensional ambient space (\eg, images or text) can be a complex process, cross-modal retrieval is often used in their place to test the goodness of the learned shared representation. Recently, recurrent neural network (RNN) architectures such as Long Short-Term Memory (LSTM) \cite{hochreiter1997} units and Gated Recurrent Units (GRU) \cite{cho2014} have (re-)emerged as effective models to capture long-term dependencies in sequential data. On the other hand, the advancement of Convolutional Neural Networks (CNN) and the wide availability of pre-trained networks, such as ResNet50~\cite{he2016}, on vast amounts of data, usually provides a strong visual backbone for joint representation learning. Most previous works on cross-modal image-to-recipe retrieval in the food domain use a pre-existing CNN backbone and treat textual elements (\eg, words) as a linear sequence in a RNN \cite{salvador2017,chen2018,carvalho2018,wang2019,Fu2020,Zhu2019, wang2020}. However, more recent advances~\cite{chef2021,salvador2021} have largely moved away from sequential RNN architectures and towards transformer models~\cite{Vaswani2017,Devlin2018} or as tree-structured network topologies such as Tree-LSTM~\cite{tai2015,zhu2015,choi2017}

In this work we propose a joint representation learning method that is able to accurately capture joint cross-modal information, and thus can effectively learn a translation between images of food and their corresponding free flowing text recipes. In order to demonstrate the validity and richness of the learned cross-modal representation, we evaluate the proposed embeddings in a cross-modal retrieval scenario using the publicly available Recipe1M (R1M) dataset, where the goal is to find matching instances across modalities (\eg, images and recipes). Results indicate the proposed method is able to substantially outperform the previous state-of-the-art (SOTA), leading to significant improvements in terms of median rank and recall\footnote{See \autoref{sec:exp} for metric definitions.} Furthermore, we propose the use of a generative model to synthesize food images conditioned on their corresponding recipe embedding, in order to assert the quality of the learned shared representation. Images generated in this manner 
achieve R1 of 0.68 in the image-to-recipe retrieval task 
and
high image fidelity ($FID < 30$), attesting to the ability of embeddings to capture the essence of the paired image-text samples. 

\begin{figure*}[!ht]
\begin{center}
\includegraphics[width=0.9\textwidth]{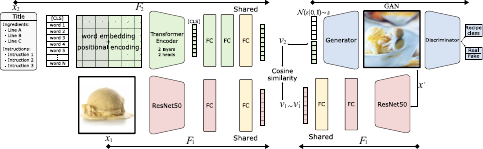}
\end{center}
\caption{Left: Cross-modal retrieval framework. Word embedding is based on word pieces from a pre-trained multilingual BERT model. Right: Synthesis network. 
}
\label{fig:diagram}
\vspace{-1em}
\end{figure*}

\section{Related Work}
Cross-modal representation learning, and particularly at the intersection between image and text representations, is an active research topic\cite{lee18,gu18,huang19}. Similarly, text induced conditional image synthesis has recently received a significant amount of attention due to its ability to ``translate'' textual representation in to high-fidelity images. Typically, text conditioning is achieved by combining a generative model such as a GAN with the encoding of a text description \cite{reed16,han2017stackgan,Xu_2018_CVPR,Zhu_2019_CVPR}. Most previous works have focused on matching and/or conditioning based on short textual captions ($\sim$11-12 words \cite{coco,flicker30k}) that describe the current state or appearance of an image. This in contrast to cooking recipes' generally much longer ($\sim$150 words \cite{salvador2017,marin2019}) descriptions of the past state of various entities (and their derivatives), that are transformed according to some actions (and utensils), defined in a set of instructions, to produce the current state depicted in the image.

\subsection{Image-Recipe Cross-Modal Representation}
Some previous works on (food) image-to-recipe retrieval have attempted to exploit this highly structured nature of textual recipes, mainly, by treating different components (title, ingredients and/or instructions) independently, before downstream merging and final latent space embedding~\cite{salvador2017,carvalho2018,chen2018,Zhu2019,wang2020,wang2019,salvador2021,Fu2020,chef2021}. However, manually enforcing recipe structure risks over (or under) representing certain components during fusion (typically through concatenation) of their individual representations. Furthermore, many of these works rely on domain specific pre-trained models for textual representations, such as word2vec~\cite{mikolov2013}, skip-thoughts~\cite{kiros2015} or an ingredient name parser\footnote{Ingredients in recipes are usually small descriptions that include quantity, unit, ingredient name, and possibly a qualifier.}, and hence add additional complexity and data requirements. Learning an embedding space that is able to capture cross-modal semantic meaning can be a challenging task, and some previous works have attempted to regularize the typical contrastive learning approach through the addition of generative~\cite{wang2019,Zhu2019} or predictive~\cite{salvador2017,carvalho2018,Zhu2019,wang2019,wang2020} loss terms. However, learning effective decoders, such as GAN or VAE, into high-dimensional ambient space (\eg, images or text), can be a complex process leading to sub-optimal latent space representations. Similarly, predictive loss terms used on previous works rely on semi-automatic labels (derived from a subset or additional data), which define either extracted ingredient names or recipe classes, and thus might not provide an effective learning signal.

The method proposed here departs from previous approaches in that we makes no intermediate structural assumptions, \eg, representing ingredients, instructions and title independently in a fixed size space before downstream fusion. This structural difference is similar to~\cite{Fain2019}, however, we further differentiate from this and other approaches in that our framework does not rely on domain specific pre-trained models, the performance of which would undoubtedly affect the quality of the embedding space. Furthermore, the proposed methodology is substantially more data efficient than the previous SOTA methods, as it does not require a large corpus of mono-modal data in order to pre-train the text encoder~\cite{Fain2019} or to provide and additional training signal~\cite{salvador2021}. Additionally, the lack of intermediate structural assumptions, naturally offers a graceful way to handle unstructured inputs and missing information (\eg a partial recipe), without the need to estimate and impute the missing information as in~\cite{salvador2021}.

Finally, and as noted before, previous approaches have attempted to regularize training through additional predictive and/or generative loss terms. Here, multilingual encoding is used in this regard as it offers significant advantages: The ability of the network to handle different language recipes, useful in a real world system, as well as sharing the semantics of different languages through the cross-modal shared latent space. Similarly, back-translation~\cite{sennrich2016}, which has been successfully used for text augmentation in different tasks~\cite{sennrich2016,xie2020}, is able to generate paraphrases while preserving semantic meaning~\cite{xie2020}, thus providing further regularization. As with any augmentation, a noise induced dampening effect is introduced as form or regularization. However, as translations are done over all recipes from different languages, noise artifacts should average out, whilst regularizing training. Empirically, this was crucial for training complex text encoders (\eg, transformer), while avoiding over-fitting.

\subsection{Cross-Modal Synthesis}
\noindent\textbf{Recipe-to-image synthesis.}
Most text-conditioned image synthesis methods recently proposed rely on short current state textual descriptions. These, generally make use of a mono-modal text encoder that captures semantic meaning of short text captions \cite{reed16,han2017stackgan,Xu_2018_CVPR,Zhu_2019_CVPR}. As mentioned before, such short captions only offer a present state description of an image and thus are far less complex than the temporal and transformative descriptions found in cooking recipes. Furthermore, mono-modal pre-trained encoders can only capture textual semantic meaning, thus ignore the intersection between modalities. This domain gap can be bridged by additionally learning or fine-tuning the text encoder (or parts of it) during the generative model training, however, due to the complexity of learning an effective decoder, this might result in sub-optimal cross-modal textual representations. Recently, cross-domain retrieval and synthesis frameworks have attempted to alleviate this, particularly, for complex cooking recipe descriptions \cite{wang2019,Zhu2019,Han2020,zhu2020}. These last methods can be spilt into joint \cite{wang2019,Zhu2019}, and separate \cite{Han2020,zhu2020}\footnote{Both \cite{Han2020,zhu2020} methods are called CookGAN, we will use CookGAN-a for \cite{Han2020} and CookGAN-b for \cite{zhu2020} to reflect order of publication.}, embedding and synthesis. The method proposed here is closely related to these methods, however, it significantly differs on how the conditional information is generated, as stated above.

\noindent\textbf{Image-to-recipe generation.}
This task is more 
complex
than regular image captioning~\cite{Vinyals2015,Karpathy2017} 
due to
the difficulty in decoding long recipe texts. Nishimura et al.~\cite{Nishimura1,Nishimura2} approached this problem by generating instructions from a sequence of images. Wang et al.~\cite{wang_eccv20} first estimate the intermediate tree-structured representation of cooking instructions from an image, and generate full sentences from it. Salvador et al.~\cite{salvador_2019} generate the full recipe including title, ingredients and instructions using a staged decoding model. Further discussion on this topic is beyond the scope of our work.

\section{Proposed Model}
In this section we introduce our proposed cross-modal retrieval model, as well as the recipe-to-image synthesis model employed in our experimental analysis. \autoref{fig:diagram} illustrates the overall architecture of the proposed cross-modal retrieval framework (left side) and food image synthesis model (right side).

\subsection{Retrieval Model}

Cross-modal learning is an active research topic in computer science. In general terms, it describes a system that, given a view or modality (\eg, image) of an instance, retrieves the same instance but being viewed in another modality (\eg, text). 
These types of models are usually trained using a direct correspondence between pairs of instances in different modalities. In the case of food recipe retrieval, these modalities are usually food images and their associated text recipe descriptions.

In order to extract feature representations from images, most previous works use either a ResNet50 or ResNext101 backbone, our architecture here makes use of ResNet50 as a visual encoder $F_1$. Here, the last fully connected (FC) layer of ResNet50 is replaced by two consecutive 1024 dimensional FC layers, the last of these being shared with the recipe encoder, that projects into the shared space.

Recipes are generally represented as three different entities: title, ingredients and instructions. Previous works handle the use and encoding of these three elements differently. Encoding ingredients usually requires a pre-trained ingredient parser plus an RNN, or two RNNs (one to encode words in an ingredient line and another to encode the sequence of lines). Encoding instructions generally make use of a pre-trained skip-thoughts model plus an RNN, or two RNNs (one to encode the sequence of words in an instruction and another for the sequence of instructions). Finally, encoding the title usually involves a single RNN. Departing from most previous works, we simplify the process and employ a single transformer encoder network \cite{Vaswani2017} in place of the complex combination of RNNs. This is achieved by treating recipes as a single long list of words, similar to \cite{Fain2019}. Hence, we make use of the complete recipe information: Title, instructions and the complete list of free flowing natural text ingredients, without any preprocessing. 

Our recipe encoder $F_2$ 
uses
WordPiece~\cite{Wu2016} to tokenize the full sequence of recipe words. Recipe words, plus an additional [CLS] token, are encoded into 768 dimensional vectors, then fed to a two-layer two-head transformer encoder model. Similar to \cite{Devlin2018}, recipes with more than 512 tokens are truncated to control the memory footprint of the model. The output of the [CLS] token is used as an aggregate recipe representation, and passed through two 1024 dimensional FC layers, before a final 1024 dimensional FC layer (shared with the visual encoder) that projects into the shared space.

The retrieval model is comprised of two encoding functions, $F_1$ and $F_2$, understood to be two neural networks of the form described above. During training, the retrieval model takes a
image-text pair $(x_1,x_2)$ as input and outputs their corresponding shared representation $(v_1,v_2)$. The model's objective is formulated as the minimization of the distance between an anchor $\anchor$ ($v_1$ or $v_2$) and a positive paired sample $\positive$, while also maximizing (up to a margin $\epsilon$) the distance between the anchor $\anchor$ and a negative non-paired sample $\negative$, that is, it minimizes the margin triplet loss of $(\anchor, \positive, \negative)$. Formally, with the image encoder $v_1 = F_1(x_1)$ and text encoder $v_2 = F_2(x_2)$, 
training minimizes the following loss function:
\begin{equation}
    L_{cos}(\anchor, \positive, \negative) =  \max\left[  d{\left(\anchor, \negative \right)} - d{\left(\anchor, \positive \right)} + \epsilon, 0 \right] ,
    \label{eq:loss}
\end{equation}

\noindent where $d\left(\cdot \right)$ is the cosine similarity.
The goal of margin $\epsilon$ is to focus the model on ``hard'' examples (negatives within the margin) while ignoring those that are beyond the margin.

\subsection{Synthesis Model}

In this work, the benefits of proposing a GAN model conditioned on recipe embedding are two-fold. Firstly, the generative model can synthesize novel images of a recipe, or visually realize any recipe without accompanying images. Secondly, we hypothesize that if the recipe embedding contains rich information of the original recipe, it should be possible to learn a generative model that can re-create an image describing the exact recipe from its embedding. In other words, besides looking realistic, the synthetic image should reflect the content of the recipe (\eg, ingredients and/or colors), and this similarity can be measured by the aforementioned retrieval model. The retrieval and synthesis models are trained independently, since our primary experiments showed that jointly training the two models as in~\cite{wang2019} led to degraded performance.

Our GAN model is a simplified version of StackGAN~\cite{Zhang2017-jm}, in which the intermediate-resolution images are discarded, and the discriminator also includes a recipe classifier, as this class information is available in the dataset. The input recipe embedding $v_2$ is expanded through a Conditional Augmentation (CA)~\cite{Zhang2017-jm} subnetwork $F_{ca}$ to create a conditioning code $t = F_{ca}(v_2)$, before passing to the image decoder $G_0$ together with the Gaussian noise $z$ to generate the fake image $x' = G(v_2,z) = G_0(t,z)$. The discriminator $D$ tries to separate 
real $x_1$ from fake $x'$ images
($D_r$ subnetwork), as well as recognize the correct recipe classes $c$ from images ($D_c$ subnetwork).
%
The discriminator's training loss is formulated as:

\begin{align}
    \begin{split}
    L_D & = \mathbb{E}_{x_1 \sim {x_1}_{data}}[logD_{r}(x_1)] + \mathbb{E}_{v_2 \sim {v_2}_{data}}[log(1 - D_r(G(v_2,z))] \\ 
    & + \lambda _c (L_c(D_c (x_1), c) + L_c (D_c (G(v_2,z)), c)),
    \label{eq:d_loss}
    \end{split}
\end{align}
where $L_c$ is the 
cross-entropy loss and $c$ 
the class label of 
$(x_1, x_2)$. The generator is learnt by minimizing the following loss function:

\begin{flalign}
    \begin{split}
        L_G & = \mathbb{E}_{v_2 \sim {v_2}_{data}}[log(1 - D_r(G(v_2,z))] 
        \\ & + \lambda _c L_c(D_c(G(v_2,z)), c) + \lambda _{ca} L_{ca} + \lambda _{ret} L_{ret},
        \label{eq:g_loss}
    \end{split}
    \\
    \begin{split}
        L_{ca} & = D_{KL}(\mathcal{N}(\mu(v_2), \Sigma (v_2)) || \mathcal{N}(0,I)),
        \label{eq:ca_loss}
    \end{split}
    \\
    \begin{split}
        L_{ret} & = \mathcal{D}(F_1(G(v_2,z)), v_2) + \mathcal{D}(F_1(G(v_2,z)), v_1),
        \label{eq:r_loss}
    \end{split}
\end{flalign}
where $L_{ca}$ is the regularization term of the CA subnetwork, $L_{ret}$ is the supervision term to enforce that the visual embedding of the fake image, $F_1(x')$, resembles those of the real image and recipe, and $\mathcal{D}(\cdot)$ is the cosine distance function. $L_{ret}$ is critical in learning to generate the ``correct'' food images. We will demonstrate its importance in our image synthesis experiments in comparison to ACME~\cite{wang2019}, which does not enforce this supervision effectively.

\section{Experiments}
\label{sec:exp}

\subsubsection*{\textbf{Implementation details.}} For the retrieval network, Adam optimizer \cite{adam} is employed for model training with an initial learning rate $10^{-4}$, which is decreased after 40 epochs to $10^{-5}$. The margin $\epsilon$ in \autoref{eq:loss} is empirically set to 0.3 by cross-validation. Batch size is set to 320, where these are understood to be formed of 320 positive pair (image, recipe) samples. Furthermore, each batch contains $2*(320-2)$ negative samples per positive pair. Training network updates are based on a loss calculated using online hard example mining, that is, at each step the loss takes only into account the hardest possible triplet per sample pair. The model is trained on 8 Nvidia GTX1080Ti graphics cards and takes approximately 20 hrs to converge. In our recipe encoder $F_2$, both WordPiece dictionary and word embedding layer are taken from the pre-trained ``bert-base-multilingual-cased'' provided in the Huggingface~\cite{huggingface} library. Furthermore, during training the word embedding weights are treated as learnable parameters. For recipe-to-image synthesis, we adopt the same generator and discriminator architectures as ACME~\cite{wang2019} for a fair and direct comparison to their work. These are a simplified version of StackGAN~\cite{Zhang2017-jm}. The generator produces single resolution synthetic images at $128^2$ pixels. The conditional supervision loss term weight, $\lambda _{ret}$, is chosen as 32 via cross-validation, while other weights are equally set to 1. Synthetic images are upsampled to $224^2$ pixels before passing to the image encoder $F_1$. The generator and discriminator networks are trained alternately for 300 epochs. Learning rate is initially set to $10^{-4}$, which is decreased to $10^{-5}$ after 30 epochs.

\subsubsection*{\textbf{Dataset.}}
Experiments presented in this work were all conducted using data from R1M \cite{salvador2017,marin2019}. This dataset consists of $\sim$1M text recipes that contain titles, instructions and ingredients in English. Additionally, a subset of $\sim$0.5M recipes contain at least one image per recipe. Data is split in 238999 train, 51119 validation and 51303 test image-recipe pairs, in accordance to the official data release provided in R1M. For the work presented here, R1M English (EN) recipes were augmented via back translation from German (EN-DE-EN) and Russian (EN-RU-EN). Additionally, all recipes were translated into German (DE), Russian (RU), French (FR) and Korean (KO). Translations from/to German, Russian and French were obtained using pre-trained models from the fairseq~\cite{ott2019fairseq} neural translation toolbox. Translation into Korean was done by an in-house system. Note that five different languages and three different alphabets (Roman, Cyrillic and Hangul) are considered here.

\subsubsection*{\textbf{Metrics.}} Retrieval models are evaluated using the same metrics as previous works, namely, median rank (medR) and recall rate at top $K$ (R$K$) are used to evaluate retrieval accuracy, with lower, and respectively higher, values indicating better performance. Here, R$K$ measures the percentage of true positives being ranked within the top $K$ returned results and inline with previous works we report values at $K=1,5,10$. Both medR and R$K$ are calculated based on a search pool of either 1k or 10k test samples, with the average over 10 different subset reported. For synthesis models, besides reporting the retrieval performance on synthetic food images, we also calculate the Fretchet Inception Distance (FID), which measures the similarity between real and synthetic image distributions. Lower FID values indicate better performance,

\begin{table*}[!ht]
  \centering
  \small
  \caption{R1M cross-modal retrieval performance comparison of our proposed methods and other reported frameworks. Methods that not require any additional mono-modal data marked with $\dagger$.}
  \begin{tabular}{l|cccc|cccc|cccc|cccc}
\toprule
	 & \multicolumn{8}{c|}{\textbf{1k}} & \multicolumn{8}{c}{\textbf{10k}}    \\ \cmidrule{2-17}
	 & \multicolumn{4}{c|}{\textbf{Image-to-Recipe}} & \multicolumn{4}{c|}{\textbf{Recipe-to-Image}}& \multicolumn{4}{c|}{\textbf{Image-to-Recipe}} & \multicolumn{4}{c}{\textbf{Recipe-to-Image}}   \\  \cmidrule{1-17}
	\textbf{Method} & \textbf{medR} & \textbf{R1} & \textbf{R5} & \textbf{R10} & \textbf{medR} & \textbf{R1} & \textbf{R5} & \textbf{R10} & \textbf{medR} & \textbf{R1} & \textbf{R5} & \textbf{R10}& \textbf{medR} & \textbf{R1} & \textbf{R5} & \textbf{R10}  \\ 
	\midrule
	Pic2recipe~\cite{salvador2017} & 5.2 & 24 & 51 & 65 & 5.1 & 25 & 52 & 65 & 41.9 & - & - & - & 39.2 & - & - & -  \\ 
	AdaMine~\cite{carvalho2018} & 2 & 39.8 & 69 & 77.4 & 2 & 40.2 & 68.6 & 78.7 & 14.2 & 14.9 & 35.3 & 45.2 & 13.2 & 14.8 & 34.6 & 46.1 \\ 
	ATTEN~\cite{chen2018} & 4.6 & 25.6 & 53.7 & 66.9 & 4.6 & 25.7 & 53.9 & 67.6 & 39.8 & 7.2 & 19.2 & 27.6 & 38.1 & 7 & 19.4 & 27.8 \\ 
	R2GAN~\cite{Zhu2019} & 2 & 39.1 & 71 & 81.7 & 2 & 40.6 & 72.6 & 83.3 & 13.9 & 13.5 & 33.5 & 44.9 & 12.6 & 14.2 & 35 & 46.8 \\ 
	ACME~\cite{wang2019} & \textbf{1} & 51.8 & 80.2 & 87.5 & \textbf{1} & 52.8 & 80.2 & 87.6 & 6.7 & 22.9 & 46.8 & 57.9 & 6 & 24.4 & 47.9 & 59.0 \\ 
	ACME* & 1.55 & 49.7 & 79.1 & 86.9 & 1.2 & 51 & 79.6 & 86.9 & 7.3 & 20.8 & 44.6 & 56 & 7 & 22.2 & 45.8 & 57.1 \\ 
	CookPad~\cite{Fain2019} & \textbf{1} & 60.2 & 84 & 89.7 & - & - & - & - & 4 & 30 & 56.5 & 67 & - & - & - & - \\ 
	$^\dagger$MCEN~\cite{Fu2020} & 2 & 48.2 & 75.8 & 83.6 & 1.9 & 48.4 & 76.1 & 83.7  & 7.2	& 20.3	& 43.3	& 54.4	& 6.6	& 21.4	& 44.3	& 55.2 \\ 
	$^\dagger$SCAN~\cite{wang2020} & 1 & 54.0 & 81.7 & 88.8 & 1 & 54.9 & 81.9 & 89.0 & 5.9 & 23.7 & 49.3 & 60.6 & 5.1 & 25.3 & 50.6 & 61.6  \\ 
	H-T~\cite{salvador2021} & \textbf{1} & 60.0 & 87.6 & \textbf{92.9} &\textbf{1} & 60.3 & 87.6 & \textbf{93.2}  & 4.0	& 27.9	& 56.4	& 68.1	& 4.0	& 28.3	& 56.5	& 68.1 \\ \hline
	$^\dagger$(Ours) B & \textbf{1} & 61.0 & 84.3 & 89.7 & \textbf{1} & 61.7 & 84.4 & 89.7 & 3.6 & 31.9 & 58.0 & 67.9 & \textbf{3} & \textbf{33.1} & 58.7 & 68.3 \\ 
	$^\dagger$(Ours) T & \textbf{1} & 60.2 & 84.6 & 89.9 & \textbf{1} & 60.7 & 84.9 & 90.0 & 4 & 30.6 & 56.7 & 67.4 & 4 & 30.9 & 57.4 & 67.8 \\ 
	$^\dagger$(Ours) B-ML & \textbf{1} & 61.0 & 84.7 & 90.5 & \textbf{1} & 60.9 & 84.9 & 90.7 & 3.9 & 31.3 & 57.5 & 67.6 & 3.8 & 32.1 & 57.5 & 67.7 \\ 
	$^\dagger$(Ours) T-ML & \textbf{1} & \textbf{64.0} & \textbf{88.3} & 92.6 & \textbf{1} & \textbf{63.9} & \textbf{87.6} & 92.6 & \textbf{3} & \textbf{32.9} & \textbf{60.6} & \textbf{71.2} & \textbf{3} & 33 & \textbf{60.4} & \textbf{70.7} \\ 
	\bottomrule
     \end{tabular}%
     
  \label{tab:xretrieval}%
 \end{table*}%

\subsection{Cross-Modal Retrieval}

In this section we compare our proposed retrieval model to several previously proposed methods, including current state-of-the-art (SOTA) methods, as well as a surprisingly strong baseline often overlooked in previous literature. As mentioned before, the proposed model's visual encoder $F_1$ makes use of a ResNet50 backbone pre-trained on ImageNet, and requires no additional pre-training on food related data, as done in \cite{Fain2019,chen2018}. Similarly, our recipe encoder $F_2$ is not pre-trained on any food related task as in \cite{Fain2019}, and does not use R1M unpaired recipes for word2vec pre-training as in \cite{salvador2017,carvalho2018,Zhu2019,wang2019} or in self-supervision tasks during training as in \cite{salvador2021}. Hence, greatly simplifying the training procedure. During training, sample pairs are randomly selected and each modality is augmented using the following criteria:
\begin{itemize}
    \item \textbf{Image augmentation.} 1) Random input image for sample recipes with more than one. 2) Randomly select, pad (with a random choice between zeros, edge duplication or image reflection) to square and resized to $256^2$, or resized to $256^2$. 3) Random rotation $\pm10^\circ$ 4) Randomly cropped to $224^2$. 5) Random horizontal flip.

    \item \textbf{Recipe augmentation.} 1) Randomly select between original EN representation and back translation from EN-DE-EN or EN-RU-EN. 2) Random selection between previous EN choice and either KO, DE, RU or FR.
\end{itemize}

\subsubsection{\textbf{Comparison to SOTA}}

In this work, we introduce two models, a baseline (B) similar to that presented in \cite{Fain2019}
and a transformer based retrieval system (T) that achieves a new SOTA by a large margin. Model B encodes recipes via a word vector average, followed by two dense layers (the second shared with the image encoder) that project into the shared latent space. It mainly differs from  \cite{Fain2019} in the use of the shared layer, tokenizer (mBERT wordPieces), and that requires no pre-training on a self-supervised target, any additional data or the pre-extraction of target labels. Both of our models are trained with and without multilingual data, we refer to these models as B-ML and and T-ML. 
Cross-modal
retrieval performance of our proposed models are compared against other works and baselines in \autoref{tab:xretrieval}. 
Note
that upon inspection of the code provided by the authors of ACME~\cite{wang2019} a bug was found\footnote{Batch normalization of the final shared layer of the model used running (as opposed to learned) statistics during testing (from Pytorch, not in evaluation mode). Batches contain query-answer pairs, and each batch is normalized by slightly different statistics, thus giving an unfair advantage. Reproducibility issues are also reported in \cite{Fain2019}.}, therefore, we report results in ACME\cite{wang2019} as well as those recalculated (using the authors' corrected code and pre-trained model), which we denote ACME*. Similarly, code made public by the authors of AdaMine indicates that their reported medR values are calculated using ranking values starting from zero, which indicates an exact match. This is in contrast to most other works where rank indices start from one. Here, medR values reported for AdaMine are shifted so that one reflects an exact match in order to allow a direct comparison to other methods\footnote{\url{https://github.com/Cadene/recipe1m.bootstrap.pytorch/blob/b009f7fee8e826693b64f3ec53ae5c8587bcd8fb/recipe1m/models/metrics/trijoint.py\#L131-L150}}. As can be seen in \autoref{tab:xretrieval}, our proposed simple baseline (B) outperforms most other previous methods, with one of the closest competitor (CookPad\footnote{No recipe-to-image results provided, therefore, only partial comparisons are possible.}) also having the most similar text encoder, a simple word vector average. Our proposed transformer model (T), despite its extra capacity and complexity is not able to surpass the very strong baseline (B). In fact, even the more complex H-T \cite{salvador2021} models, which is based on hierarchical transformers and uses vast amounts of mono-modal data for self-supervision, strugles to compete with baseline (B). Additionally, it can be observed that the baseline (B-ML) does not degrade in performance when adding multilingual capabilities. On the other hand, the proposed transformer based model (T-ML) significantly improves over the baseline and SOTA. The results from our proposed model T-ML improved most metrics on both 1K and 10K search space experiments.  
Overall, between B-ML and T-ML, the retrieval rank of $\sim$6k samples is improved, this represents $\sim$12\% of the test data. \autoref{fig:i2r_r2i} shows some image-to-recipe an recipe-to-image qualitative ``multilingual'' retrieval results using the embeddings of our  
T-ML model from \autoref{tab:xretrieval}. Note that since our recipe embedding network is based on transformer encoders, we can implicitly obtain attention over the different words in a recipe. WordPiece attention is visualized and calculated as proposed in \cite{Abnar2020}. Additionally, in \autoref{fig:i2r_r2i} it can be observed that the recipe encoder focuses mainly on the ingredients section, and in particular mostly on those words that 
help describe the visual appearance 
of the retrieved samples.

\subsubsection{\textbf{Ablation Study}}

In order to test the amount of information contributed by title, ingredient lines and instructions toward retrieval performance, an ablation analysis was carried out. 
\autoref{tab:ablation} gives input ablation results, as well as a comparison to \cite{salvador2021} on this same inputs. 

\begin{figure}
\centering
\includegraphics[width=\linewidth]{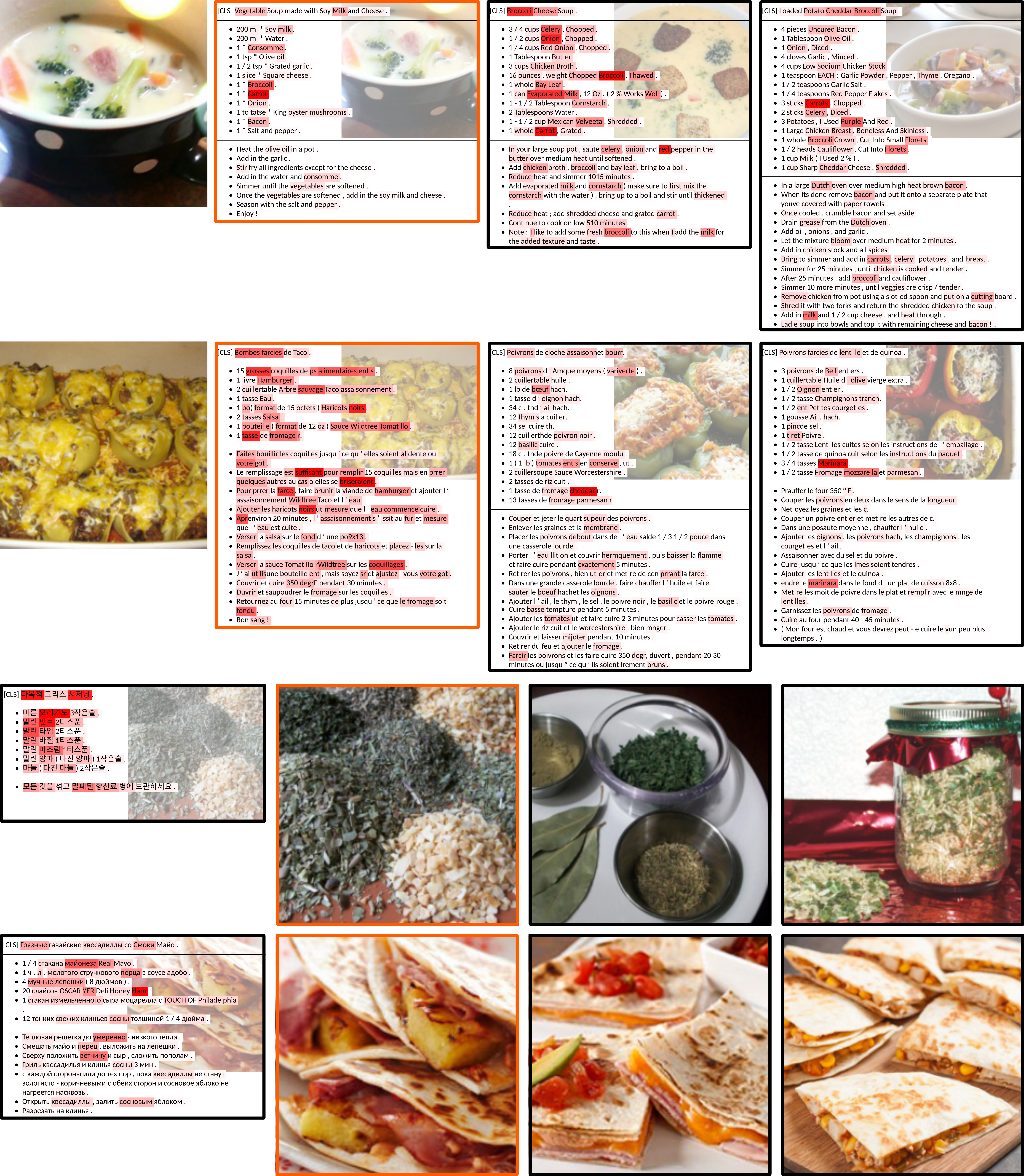}
\caption{Qualitative retrieval examples. From left to right, query and top-3 retrievals. Notice that attention over recipes primarily focuses on ingredients and words that help describe the visual appearance of retrieved samples./media/r.guerrero/DataDrive/PycharmProjects/swav/runs}
\label{fig:i2r_r2i}
\vspace{-1em}
\end{figure}

\autoref{tab:ablation} indicates that, when retrained on ingredients alone or model achieves $\sim$80\% of full recipe performance. This observation is further highlighted in the attention visualizations found in \autoref{fig:i2r_r2i}. While results reported in \cite{salvador2021} largely agree with these findings, these results are in contrast to \cite{chen2018}, where it was reported that ingredients alone offered very poor performance. However, it is important to note that ingredients used in \cite{chen2018} refer to a list of pre-extracted ingredient name tokens, whereas in our analysis and in \cite{salvador2021} ingredients refers to the full 
natural text found in recipes, thus, the ingredients set contains far more information in our analysis. As expected, the use of the two strongest elements (i.e. instructions and ingredients) performs better than using weaker elements (\eg, title and ingredients). Nonetheless, the last row of \autoref{tab:ablation} clearly indicates that there is value in using all pieces of information.
Results reported in \cite{salvador2021} largely agree with these findings.

On close inspection of \autoref{tab:ablation}, it can be observed that proposed T-ML compares very favourably against H-T \cite{salvador2021} when both methods are trained on incomplete data. In this scenario, T-ML always outperforms \cite{salvador2021} by a margin. However, evaluating partial data input, on models trained on complete recipes, it can be observed that \cite{salvador2021} performs best when the input has few words (\eg title), while T-ML performs best when input sequences approach a similar length to those of complete recipes (\eg ingredients plus instructions).

\begin{table}[!ht]
  \centering
  \small
  \caption{Ablation and comparison 
  of textual recipe components. Performance metrics reported for 10k search space to allow comparison to H-T\cite{salvador2021}. ``Tr'' indicates whether the model was trained using only partial inputs. Note that H-T\cite{salvador2021} uses additional unpaired recipe data during training.}
  \begin{tabular}{l|l|l|cccc}
\toprule
	 \multicolumn{7}{c}{\textbf{Image-to-Recipe Retrieval}} \\ \hline
	\textbf{Input} & \textbf{Tr} & \textbf{Method} & \textbf{medR} & \textbf{R1} & \textbf{R5} & \textbf{R10} \\ 
	\midrule
	\multirow{4}*{title} & \multirow{2}*{Yes}  & (Ours) T-ML     & \textbf{32.3} & \textbf{8.8} & \textbf{23.2} & \textbf{32.6} \\ 
	                     & & H-T~\cite{salvador2021}   & 35.5 & 6.0 & 18.9 & 28.4 \\ \cmidrule{2-7}
	                     & \multirow{2}*{No} & (Ours) T-ML       & 85.6 & 5.6 & 15.0 & 21.9 \\ 
	                     & & H-T~\cite{salvador2021}   & \textbf{35.8} & \textbf{6.6} & \textbf{20.0} & \textbf{29.3} \\ \cmidrule{1-7}
	                     
	\multirow{4}*{ingr} & \multirow{2}*{Yes}  & (Ours) T-ML      & \textbf{6.0} & \textbf{25.1} & \textbf{49.4} & \textbf{60.0} \\ 
	                     & & H-T~\cite{salvador2021}   & 8.3 & 19.2 & 42.5 & 53.9 \\ \cmidrule{2-7}
	                     & \multirow{2}*{No} & (Ours) T-ML       & 11.4 & 17.7 & 38.6 & 48.8 \\ 
	                     & & H-T~\cite{salvador2021}   & \textbf{8.0} & \textbf{19.4} & \textbf{43.5} & \textbf{55.3}
 \\ \cmidrule{1-7}
	                     
	\multirow{4}*{inst} & \multirow{2}*{Yes}  & (Ours) T-ML      & \textbf{7.3} & \textbf{21.7} & \textbf{44.7} & \textbf{55.8} \\ 
	                     & & H-T~\cite{salvador2021}   & 15.0 & 13.1 & 32.6 & 43.8 \\ \cmidrule{2-7}
	                     & \multirow{2}*{No} & (Ours) T-ML       & 14.0 & \textbf{15.6} & \textbf{35.2} & 45.3 \\ 
	                     & & H-T~\cite{salvador2021}   & \textbf{13.9} & 14.0 & 34.1 & \textbf{45.4} \\ \cmidrule{1-7}
	                     
	\multirow{4}*{\begin{tabular}{@{}c@{}}title, \\ ingr\end{tabular}} & \multirow{2}*{Yes} & (Ours) T-ML & \textbf{4.0} & \textbf{30.6} & \textbf{57.6} & \textbf{68.2} \\ 
	                     & & H-T~\cite{salvador2021}   & 6.0 & 22.3 & 48.0 & 59.8 \\ \cmidrule{2-7}
	                     & \multirow{2}*{No} & (Ours) T-ML       & \textbf{5.4} & \textbf{24.7} & \textbf{50.0} & 60.7 \\ 
	                     & & H-T~\cite{salvador2021}   & 6.0 & 23.1 & 49.4 & \textbf{61.1} \\ \cmidrule{1-7}
	                     
	\multirow{4}*{\begin{tabular}{@{}c@{}}title, \\ inst\end{tabular}} & \multirow{2}*{Yes} & (Ours) T-ML & \textbf{5.1} & \textbf{25.1} & \textbf{50.6} & \textbf{62.0} \\ 
	                     & & H-T~\cite{salvador2021}   & 10.2 & 16.0 & 38.3 & 50.2 \\ \cmidrule{2-7}
	                     & \multirow{2}*{No} & (Ours) T-ML       & \textbf{8.0} & \textbf{20.4} & \textbf{43.7} & \textbf{55.0} \\ 
	                     & & H-T~\cite{salvador2021}   & 10.1 & 16.6 & 39.1 & 50.8 \\ \cmidrule{1-7}
	                     
	\multirow{4}*{\begin{tabular}{@{}c@{}}ingr, \\ inst\end{tabular}} & \multirow{2}*{Yes} & (Ours) T-ML  & \textbf{3.1} & \textbf{31.6} & \textbf{59.1} & \textbf{69.8} \\ 
	                     & & H-T~\cite{salvador2021}   & 6.0 & 22.7 & 48.4 & 60.4 \\ \cmidrule{2-7}
	                     & \multirow{2}*{No} & (Ours) T-ML       & \textbf{4.0} & \textbf{30.3} & \textbf{57.8} & \textbf{68.5} \\ 
	                     & & H-T~\cite{salvador2021}   & 5.0 & 24.2 & 51.2 & 63.1 \\ 

	\bottomrule
     \end{tabular}%
  \label{tab:ablation}%
 \end{table}%

\begin{table}[!ht]
  \centering
  \small
  \caption{Multi-lingual performance 
  of translated R1M.}
  \begin{tabular}{l|cccc|cccc}
\toprule
	\textbf{1k} & \multicolumn{4}{c|}{\textbf{Image-to-Recipe}} & \multicolumn{4}{c}{\textbf{Recipe-to-Image}}  \\ \hline
	\textbf{Lang.} & \textbf{medR} & \textbf{R1} & \textbf{R5} & \textbf{R10} & \textbf{medR} & \textbf{R1} & \textbf{R5} & \textbf{R10}  \\ 
	\midrule
	KO & 1.1 & 52.8 & 79.4 & 85.8 & 1 & 53.8 & 79.6 & 86.4\\ 
	DE & 1 & 57.7 & 82.8 & 88.9 & 1 & 58.0 & 83.1 & 89.2 \\ 
	RU & 1 & 53.3 & 79.6 & 86.7 & 1 & 53.7 & 80.1 & 87.0 \\ 
	FR & 1 & 57.2 & 82.5 & 88.8 & 1 & 57.5 & 82.5 & 89.1 \\ 
	\bottomrule
     \end{tabular}%
  \label{tab:multilanguage}%
 \end{table}%

\subsubsection{\textbf{Multi-lingual Retrieval Analysis}}

As previously noted, our proposed model is able to encode text in multiple languages. \autoref{tab:multilanguage} shows the retrieval performance metrics of our proposed T-ML model for the four additional languages considered here. Note that the retrieval results for EN are those found in \autoref{tab:xretrieval}. As can be seen in \autoref{tab:multilanguage} languages that use the Roman alphabet (DE and FR) perform clearly better than those that use either Cyrillic or Hangul (RU and KO). We hypothesize this is primarily due to the fact that three of the five languages considered here use the Roman alphabet, and share many entries in the WordPiece dictionary, therefore, also sharing the weights associated to them and thus possibly being regularized. 

\begin{table*}[!ht]
  \centering
  \small
  \caption{Recipe-to-image synthesis performance. We evaluate the quality of images generated from recipe embeddings by two metrics: FID of synthetic images, and retrieval scores averaged over 10 random 1K subsets. 
  Food
  images generated from CHEF \cite{chef2021} recipe embeddings generally look more realistic, as FID indicates. However, our proposed recipe embeddings can generate images more suitable for retrieval. $\dagger$ indicates values taken directly from \cite{zhu2020} as no code was available to replicate results.}
    \begin{tabular}{l|c|cccc|cccc}
    \toprule
    \multicolumn{1}{c|}{\multirow{2}[4]{*}{\textbf{Recipe Embedding}}} & \multirow{2}[4]{*}{\textbf{FID} } & \multicolumn{4}{c|}{\textbf{Synthetic Image-To-Recipe}} & \multicolumn{4}{c}{\textbf{Recipe-To-Synthetic Image}} \\
\cmidrule{3-10}           &        & \textbf{medR}    & \textbf{R1}   & \textbf{R5} $\uparrow$   & \textbf{R10}  & \textbf{medR}   & \textbf{R1}     & \textbf{R5} $\uparrow$    & \textbf{R10}  \\
    \midrule
    ACME~\cite{wang2019}   & 30.7   & 135.8  & 1.7    & 6.2    & 10.5   & 84.9   & 2.5    & 8.9    & 14.6 \\
    $\dagger$R2GAN~\cite{Zhu2019} & - & 500.0  & -    & -    & -    & -  & -    & -    & - \\
    $\dagger$CookGAN-b~\cite{zhu2020} & - & 64.0  & -    & -    & -    & -  & -    & -    & - \\
    CHEF~\cite{chef2021}   & \textbf{23.0} & 1.0    & 54.2   & 86.7   & 92.4   & 1.0    & 60.9   & 90.1   & 94.8 \\
    (Ours) T & 36.3   & 1.0    & 63.2   & 87.4   & 92.7   & 1.0    & 71.6   & 92.6   & 96.0 \\
    (Ours) T-ML & 28.6   & \textbf{1.0} & \textbf{68.0} & \textbf{91.9} & \textbf{95.8} & \textbf{1.0} & \textbf{73.1} & \textbf{94.6} & \textbf{97.3} \\
    \bottomrule
    \end{tabular}%
  \label{tab:r2i_syn}%
\end{table*}%

\subsection{Cross-Modal Synthesis}

\subsubsection{\textbf{Recipe-to-image Synthesis.}}
 
As the goal of the generator $G$ is to synthesize realistic food images that resemble real recipes, we evaluate two aspects of our proposed recipe-to-image synthesis model: image realism and food-likeness. The first is indicated by FID scores, while the second is reflected in retrieval performance, i.e. retrieving the correct recipe given a synthetic image. We compare these metrics of our proposed models trained on English-only transformer-based embeddings (our T model) and multilingual transformer based embeddings (T-ML model) with ACME~\cite{wang2019}, as we use very similar model architectures to their work, and trained on the same data. Moreover, we also train and evaluate our GAN model on the recipe embeddings created by the hierarchical text encoder recently proposed in CHEF~\cite{chef2021}\footnote{CHEF is trained on a modified R1M dataset, hence its retrieval performance is not included in \autoref{tab:xretrieval}.}. Additionally, we include the performance of R2GAN~\cite{Zhu2019} and CookGAN-b~\cite{zhu2020} as reported in ~\cite{zhu2020}\footnote{Code is not available to reproduce the results of these two methods.}. Note that R2GAN generates images at $64^2$ pixels, while CookGAN-b generate higher resolution images of $256^2$ pixels, hence, caution must be taken on comparisons. The results are summarized in \autoref{tab:r2i_syn}. Retrieval and FID results reported in CookGAN-a~\cite{Han2020} are not directly comparable with those reported here, as are conditioned solely on ingredients and focus only on three food classes (salad, muffin and cookie). Nonetheless, reported median FID and medR for those three classes are 81.13 and 103.30, respectively.

\begin{figure*}[!ht]
    \centering
    \includegraphics[width=\textwidth]{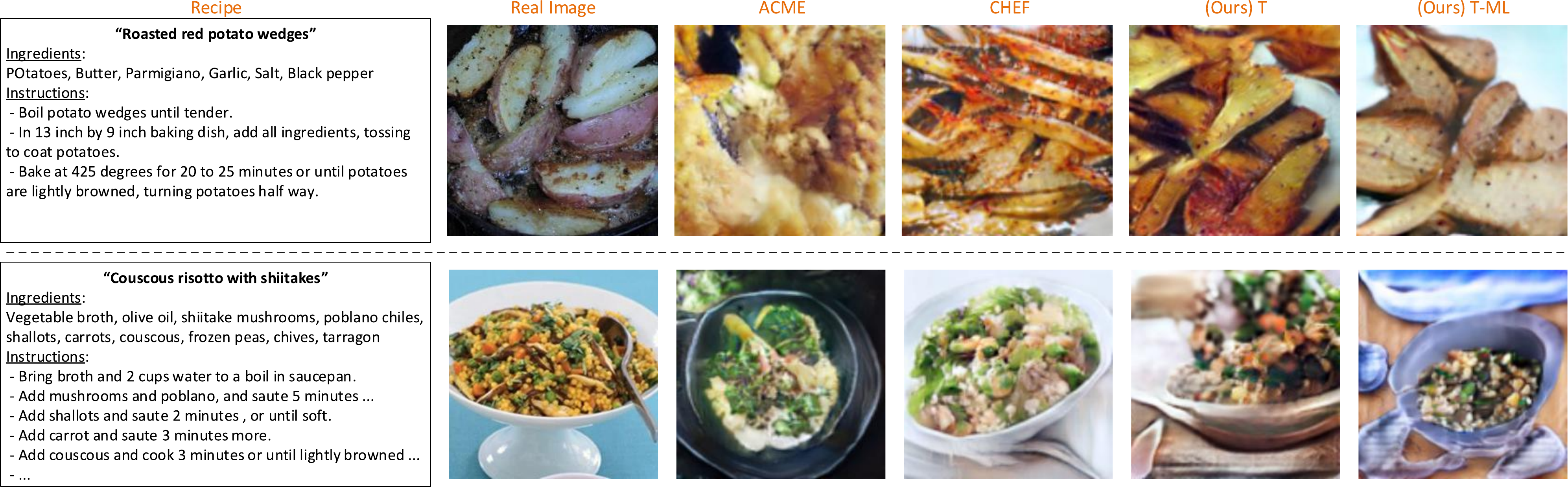}
    \caption{Recipe-to-image synthesis examples. These examples 
    are chosen such that the recipes are 
    retrieved successfully using our proposed T-ML 
    retrieval 
    model. In each row, the recipe and its associated real image are shown on the left hand side, and the next four images are generated from ACME, CHEF, (our) T and T-ML recipe embeddings, respectively. 
    }
    \label{fig:synthesis}
\end{figure*}

\begin{table*}[!ht]
  \centering
  \small
  \caption{Image embedding-to-image synthesis performance. 
  Results
  are calculated similarly to those in \autoref{tab:r2i_syn}, however, images are generated from visual embeddings instead of recipe embeddings. $\dagger$ indicates values taken directly from \cite{zhu2020}.}
    \begin{tabular}{l|c|cccc|cccc}
    \toprule
    \multicolumn{1}{c|}{\multirow{2}[4]{*}{\textbf{Recipe Embedding}}} & \multirow{2}[4]{*}{\textbf{FID} } & \multicolumn{4}{c|}{\textbf{Synthetic Image-To-Recipe}} & \multicolumn{4}{c}{\textbf{Recipe-To-Synthetic Image}} \\
\cmidrule{3-10}           &        & \textbf{medR}   & \textbf{R1}   & \textbf{R5}    & \textbf{R10}  & \textbf{medR}   & \textbf{R1}    & \textbf{R5}    & \textbf{R10}  \\
    \midrule
    ACME~\cite{wang2019} & 30.5 & 219.0  & 0.8    & 3.8    & 6.1    & 161.0  & 1.5    & 5.6    & 9.1 \\
    $\dagger$R2GAN~\cite{Zhu2019} & - & 476.0  & -    & -    & -    & -  & -    & -    & - \\
    $\dagger$CookGAN-b~\cite{zhu2020} & - & 108.0  & -    & -    & -    & -  & -    & -    & - \\
    CHEF~\cite{chef2021} & \textbf{23.4}  & 5.9    & 20.7   & 48.7   & 60.0   & 4.9    & 23.6   & 52.0   & 62.6 \\
    (Ours) T & 35.5 & 5.6    & 23.3   & 49.6   & 60.5   & 4.2    & 27.9   & 54.8   & 65.3 \\
    (Ours) T-ML & 27.9 & \textbf{3.9} & \textbf{28.5} & \textbf{56.9} & \textbf{68.0} & \textbf{3.0} & \textbf{32.5} & \textbf{60.5} & \textbf{70.7} \\
    \bottomrule
    \end{tabular}%
  \label{tab:i2i_syn}%
\end{table*}%

\begin{figure*}[!ht]
\centering
\includegraphics[width=\textwidth]{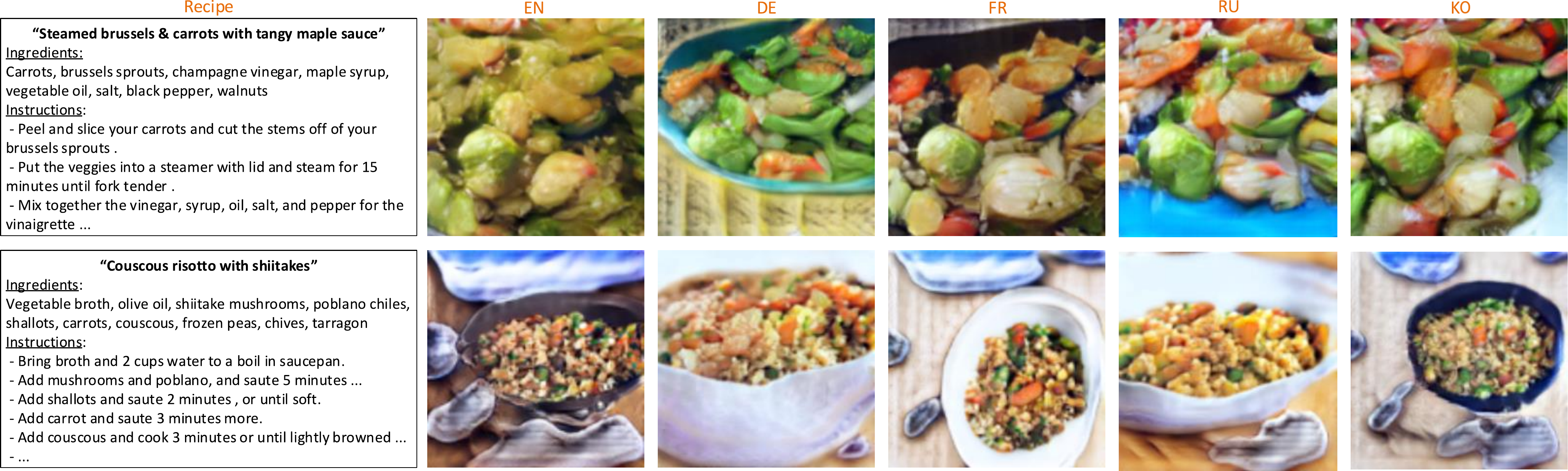}
\caption{Image synthesis from the same recipe but in different languages, using our T-ML embeddings. The original recipe (left) is written in English, and translated to different languages. In each row, (from left to right) five images are generated from the English recipe (EN), German (DE), French (FR), Russian (RU) and Korean (KO), respectively. The recipe embeddings from different languages of the same recipe can generate similarly looking food images.}
\label{fig:syn_multi_lang}
\end{figure*}

In terms of visual realism, CHEF embeddings enable the generative model to produce the most realistic looking food images among the tested methods, reflected by its smallest FID score. However, the transformer-based embeddings allow the generative model to generate images more suitable for cross-modal retrieval, evidenced by higher recall rates. In particular, synthetic images generated by the GAN model using multilingual embedding T-ML achieve very high retrieval performance, substantially better than images generated from English-only embedding in the image-to-recipe retrieval task. Furthermore, the synthetic images contain information particularly tailored for retrieval, thus recall rates on these samples are even higher than recall rates on real images shown in \autoref{tab:xretrieval}. Lastly, our proposed recipe embeddings outperform ACME, R2GAN and CookGAN-b significantly in synthetic image retrieval. A few synthesis examples are shown in \autoref{fig:synthesis} with comparison to the baseline methods. \autoref{fig:syn_multi_lang} shows five synthetic images generated from each of the ``multilingual'' (translation) embeddings, for three recipes. In each example, the images generated from five different language embeddings have similar appearance, in terms of food attributes (color, ingredients). More importantly, the generator was trained on only English recipes. These results indicate that our proposed T-ML model has properly learned the association of recipes across languages.

\subsubsection{\textbf{Evaluation of Embedding Purity}}

In the previous section, it was demonstrated that images synthesized from recipe embeddings are very suitable for image-to-recipe retrieval. However, since the generated image fully contains the recipe embedding information, it is easier to achieve high retrieval performance. Theoretically, the more effective a retrieval model is, the closer the recipe embeddings and their corresponding image embeddings are in the shared representation. In order to inspect the similarity of two modality embeddings, we conduct synthesis experiments where, instead of passing the recipe embeddings $v_2$ to the generator $G$ (as it was trained), the corresponding image embeddings $v_1$ are used. FID and retrieval scores of these visual embedding-generated synthetic images are presented in \autoref{tab:i2i_syn}.

It is observed that FID scores of synthetics images generated from four different image embeddings are roughly similar to those estimated from recipe-generated synthetic images. However, the cross-modal retrieval performance drastically degrades. Specifically, the images generated by CHEF/T/T-ML-guided visual embeddings have medR of 5.9, 5.6 and 3.9, respectively, compared to medR of 1.0 using images generated from recipe embeddings. Recall rate of CHEF drops by 33.5 points, whereas for both transformer-based models recall rates dropped by $\sim$40 points. Nevertheless, the visual embeddings guided by T-ML generate images that achieve the best retrieval performance, indicating that the two modalities of this model are aligned more closely in the common representation.

\section{Conclusion}

In this work, we present a novel multilingual cross-modal recipe retrieval framework that learns to jointly align the latent representations of images and (text) recipes in five different languages that use three different alphabets. The proposed methods achieve a new SOTA by a large margin. This is done at the same time simplifying both architecture and training procedure, as well as using less data and/or labels than other methods. Additionally, we propose a GAN framework that effectively learns to synthesize realistic food images from embeddings in the shared representation. In future work, we will explore the possibility of jointly learning both the retrieval and GAN models in an end-to-end framework.

\bibliographystyle{ACM-Reference-Format}
\bibliography{sample-base}


\begin{thebibliography}{48}


\ifx \showCODEN    \undefined \def \showCODEN     #1{\unskip}     \fi
\ifx \showDOI      \undefined \def \showDOI       #1{#1}\fi
\ifx \showISBNx    \undefined \def \showISBNx     #1{\unskip}     \fi
\ifx \showISBNxiii \undefined \def \showISBNxiii  #1{\unskip}     \fi
\ifx \showISSN     \undefined \def \showISSN      #1{\unskip}     \fi
\ifx \showLCCN     \undefined \def \showLCCN      #1{\unskip}     \fi
\ifx \shownote     \undefined \def \shownote      #1{#1}          \fi
\ifx \showarticletitle \undefined \def \showarticletitle #1{#1}   \fi
\ifx \showURL      \undefined \def \showURL       {\relax}        \fi
\providecommand\bibfield[2]{#2}
\providecommand\bibinfo[2]{#2}
\providecommand\natexlab[1]{#1}
\providecommand\showeprint[2][]{arXiv:#2}

\bibitem[\protect\citeauthoryear{Abnar and Zuidema}{Abnar and Zuidema}{2020}]%
        {Abnar2020}
\bibfield{author}{\bibinfo{person}{Samira Abnar} {and} \bibinfo{person}{Willem
  Zuidema}.} \bibinfo{year}{2020}\natexlab{}.
\newblock \bibinfo{title}{Quantifying Attention Flow in Transformers}.
\newblock
\newblock
\showeprint[arxiv]{2005.00928}~[cs.LG]


\bibitem[\protect\citeauthoryear{Carvalho, Cad\`{e}ne, Picard, Soulier, Thome,
  and Cord}{Carvalho et~al\mbox{.}}{2018}]%
        {carvalho2018}
\bibfield{author}{\bibinfo{person}{Micael Carvalho}, \bibinfo{person}{R\'{e}mi
  Cad\`{e}ne}, \bibinfo{person}{David Picard}, \bibinfo{person}{Laure Soulier},
  \bibinfo{person}{Nicolas Thome}, {and} \bibinfo{person}{Matthieu Cord}.}
  \bibinfo{year}{2018}\natexlab{}.
\newblock \showarticletitle{Cross-Modal Retrieval in the Cooking Context:
  Learning Semantic Text-Image Embeddings}. In \bibinfo{booktitle}{\emph{The
  41st International ACM SIGIR Conference on Research \& Development in
  Information Retrieval}} (Ann Arbor, MI, USA) \emph{(\bibinfo{series}{SIGIR
  '18})}. \bibinfo{publisher}{Association for Computing Machinery},
  \bibinfo{address}{New York, NY, USA}, \bibinfo{pages}{35–44}.
\newblock
\showISBNx{9781450356572}
\urldef\tempurl%
\url{https://doi.org/10.1145/3209978.3210036}
\showDOI{\tempurl}


\bibitem[\protect\citeauthoryear{Chen, Ngo, Feng, and Chua}{Chen
  et~al\mbox{.}}{2018}]%
        {chen2018}
\bibfield{author}{\bibinfo{person}{Jing-Jing Chen}, \bibinfo{person}{Chong-Wah
  Ngo}, \bibinfo{person}{Fu-Li Feng}, {and} \bibinfo{person}{Tat-Seng Chua}.}
  \bibinfo{year}{2018}\natexlab{}.
\newblock \showarticletitle{Deep Understanding of Cooking Procedure for
  Cross-modal Recipe Retrieval}. In \bibinfo{booktitle}{\emph{2018 ACM
  Multimedia Conference on Multimedia Conference}}. ACM,
  \bibinfo{pages}{1020--1028}.
\newblock


\bibitem[\protect\citeauthoryear{Cho, {Van Merri{\"{e}}nboer}, Gulcehre,
  Bahdanau, Bougares, Schwenk, and Bengio}{Cho et~al\mbox{.}}{2014}]%
        {cho2014}
\bibfield{author}{\bibinfo{person}{Kyunghyun Cho}, \bibinfo{person}{Bart {Van
  Merri{\"{e}}nboer}}, \bibinfo{person}{Caglar Gulcehre},
  \bibinfo{person}{Dzmitry Bahdanau}, \bibinfo{person}{Fethi Bougares},
  \bibinfo{person}{Holger Schwenk}, {and} \bibinfo{person}{Yoshua Bengio}.}
  \bibinfo{year}{2014}\natexlab{}.
\newblock \showarticletitle{{Learning phrase representations using RNN
  encoder-decoder for statistical machine translation}}. In
  \bibinfo{booktitle}{\emph{Conference on Empirical Methods in Natural Language
  Processing, Proceedings of the Conference}}. \bibinfo{publisher}{Association
  for Computational Linguistics (ACL)}, \bibinfo{pages}{1724--1734}.
\newblock


\bibitem[\protect\citeauthoryear{Choi, {Min Yoo}, and Lee}{Choi
  et~al\mbox{.}}{2017}]%
        {choi2017}
\bibfield{author}{\bibinfo{person}{Jihun Choi}, \bibinfo{person}{Kang {Min
  Yoo}}, {and} \bibinfo{person}{Sang-goo Lee}.}
  \bibinfo{year}{2017}\natexlab{}.
\newblock \showarticletitle{{Learning to Compose Task-Specific Tree
  Structures}}. In \bibinfo{booktitle}{\emph{In Association for the Advancement
  of Artificial Intelligence}}.
\newblock
\urldef\tempurl%
\url{www.aaai.org}
\showURL{%
\tempurl}


\bibitem[\protect\citeauthoryear{Devlin, Chang, Lee, and Toutanova}{Devlin
  et~al\mbox{.}}{2018}]%
        {Devlin2018}
\bibfield{author}{\bibinfo{person}{Jacob Devlin}, \bibinfo{person}{Ming{-}Wei
  Chang}, \bibinfo{person}{Kenton Lee}, {and} \bibinfo{person}{Kristina
  Toutanova}.} \bibinfo{year}{2018}\natexlab{}.
\newblock \showarticletitle{{BERT:} Pre-training of Deep Bidirectional
  Transformers for Language Understanding}.
\newblock \bibinfo{journal}{\emph{CoRR}}  \bibinfo{volume}{abs/1810.04805}
  (\bibinfo{year}{2018}).
\newblock
\showeprint[arxiv]{1810.04805}
\urldef\tempurl%
\url{http://arxiv.org/abs/1810.04805}
\showURL{%
\tempurl}


\bibitem[\protect\citeauthoryear{Fain, Ponikar, Fox, and Bollegala}{Fain
  et~al\mbox{.}}{2019}]%
        {Fain2019}
\bibfield{author}{\bibinfo{person}{Mikhail Fain}, \bibinfo{person}{Andrey
  Ponikar}, \bibinfo{person}{Ryan Fox}, {and} \bibinfo{person}{Danushka
  Bollegala}.} \bibinfo{year}{2019}\natexlab{}.
\newblock \showarticletitle{Dividing and Conquering Cross-Modal Recipe
  Retrieval: from Nearest Neighbours Baselines to SoTA}.
\newblock \bibinfo{journal}{\emph{CoRR}}  \bibinfo{volume}{abs/1911.12763}
  (\bibinfo{year}{2019}).
\newblock
\showeprint[arxiv]{1911.12763}
\urldef\tempurl%
\url{http://arxiv.org/abs/1911.12763}
\showURL{%
\tempurl}


\bibitem[\protect\citeauthoryear{Fu, Wu, Liu, and Sun}{Fu
  et~al\mbox{.}}{2020}]%
        {Fu2020}
\bibfield{author}{\bibinfo{person}{Han Fu}, \bibinfo{person}{Rui Wu},
  \bibinfo{person}{Chenghao Liu}, {and} \bibinfo{person}{Jianling Sun}.}
  \bibinfo{year}{2020}\natexlab{}.
\newblock \showarticletitle{MCEN: Bridging Cross-Modal Gap between Cooking
  Recipes and Dish Images with Latent Variable Model}. In
  \bibinfo{booktitle}{\emph{Proceedings of the IEEE/CVF Conference on Computer
  Vision and Pattern Recognition (CVPR)}}.
\newblock


\bibitem[\protect\citeauthoryear{{Gu}, {Cai}, {Joty}, {Niu}, and {Wang}}{{Gu}
  et~al\mbox{.}}{2018}]%
        {gu18}
\bibfield{author}{\bibinfo{person}{J. {Gu}}, \bibinfo{person}{J. {Cai}},
  \bibinfo{person}{S. {Joty}}, \bibinfo{person}{L. {Niu}}, {and}
  \bibinfo{person}{G. {Wang}}.} \bibinfo{year}{2018}\natexlab{}.
\newblock \showarticletitle{Look, Imagine and Match: Improving Textual-Visual
  Cross-Modal Retrieval with Generative Models}. In
  \bibinfo{booktitle}{\emph{2018 IEEE/CVF Conference on Computer Vision and
  Pattern Recognition (CVPR)}}. \bibinfo{pages}{7181--7189}.
\newblock


\bibitem[\protect\citeauthoryear{Han, Guerrero, and Pavlovic}{Han
  et~al\mbox{.}}{2020}]%
        {Han2020}
\bibfield{author}{\bibinfo{person}{F. Han}, \bibinfo{person}{R. Guerrero},
  {and} \bibinfo{person}{V. Pavlovic}.} \bibinfo{year}{2020}\natexlab{}.
\newblock \showarticletitle{{CookGAN}: Meal Image Synthesis from Ingredients}.
  In \bibinfo{booktitle}{\emph{WACV}}.
\newblock


\bibitem[\protect\citeauthoryear{He, Zhang, Ren, and Sun}{He
  et~al\mbox{.}}{2016}]%
        {he2016}
\bibfield{author}{\bibinfo{person}{Kaiming He}, \bibinfo{person}{Xiangyu
  Zhang}, \bibinfo{person}{Shaoqing Ren}, {and} \bibinfo{person}{Jian Sun}.}
  \bibinfo{year}{2016}\natexlab{}.
\newblock \showarticletitle{Deep residual learning for image recognition}. In
  \bibinfo{booktitle}{\emph{Proceedings of the IEEE conference on computer
  vision and pattern recognition}}. \bibinfo{pages}{770--778}.
\newblock


\bibitem[\protect\citeauthoryear{Helmy, Al-Nazer, Al-Bukhitan, and Iqbal}{Helmy
  et~al\mbox{.}}{2015}]%
        {helmy2015health}
\bibfield{author}{\bibinfo{person}{Tarek Helmy}, \bibinfo{person}{Ahmed
  Al-Nazer}, \bibinfo{person}{Saeed Al-Bukhitan}, {and} \bibinfo{person}{Ali
  Iqbal}.} \bibinfo{year}{2015}\natexlab{}.
\newblock \showarticletitle{Health, food and user's profile ontologies for
  personalized information retrieval}.
\newblock \bibinfo{journal}{\emph{Procedia Computer Science}}
  \bibinfo{volume}{52} (\bibinfo{year}{2015}), \bibinfo{pages}{1071--1076}.
\newblock


\bibitem[\protect\citeauthoryear{Hochreiter and Schmidhuber}{Hochreiter and
  Schmidhuber}{1997}]%
        {hochreiter1997}
\bibfield{author}{\bibinfo{person}{Sepp Hochreiter} {and}
  \bibinfo{person}{J{\"{u}}rgen Schmidhuber}.} \bibinfo{year}{1997}\natexlab{}.
\newblock \showarticletitle{{Long Short-Term Memory}}.
\newblock \bibinfo{journal}{\emph{Neural Computation}} \bibinfo{volume}{9},
  \bibinfo{number}{8} (\bibinfo{date}{nov} \bibinfo{year}{1997}),
  \bibinfo{pages}{1735--1780}.
\newblock
\showISSN{08997667}
\urldef\tempurl%
\url{https://doi.org/10.1162/neco.1997.9.8.1735}
\showDOI{\tempurl}


\bibitem[\protect\citeauthoryear{Huang and Wang}{Huang and Wang}{2019}]%
        {huang19}
\bibfield{author}{\bibinfo{person}{Yan Huang} {and} \bibinfo{person}{Liang
  Wang}.} \bibinfo{year}{2019}\natexlab{}.
\newblock \showarticletitle{ACMM: Aligned Cross-Modal Memory for Few-Shot Image
  and Sentence Matching}. In \bibinfo{booktitle}{\emph{Proceedings of the
  IEEE/CVF International Conference on Computer Vision (ICCV)}}.
\newblock


\bibitem[\protect\citeauthoryear{Karpathy and Fei-Fei}{Karpathy and
  Fei-Fei}{2017}]%
        {Karpathy2017}
\bibfield{author}{\bibinfo{person}{Andrej Karpathy} {and} \bibinfo{person}{Li
  Fei-Fei}.} \bibinfo{year}{2017}\natexlab{}.
\newblock \showarticletitle{Deep Visual-Semantic Alignments for Generating
  Image Descriptions}.
\newblock \bibinfo{journal}{\emph{IEEE Transactions on Pattern Analysis and
  Machine Intelligence}} \bibinfo{volume}{39}, \bibinfo{number}{4}
  (\bibinfo{year}{2017}), \bibinfo{pages}{664--676}.
\newblock
\urldef\tempurl%
\url{https://doi.org/10.1109/TPAMI.2016.2598339}
\showDOI{\tempurl}


\bibitem[\protect\citeauthoryear{Kingma and Ba}{Kingma and Ba}{2015}]%
        {adam}
\bibfield{author}{\bibinfo{person}{Diederik~P. Kingma} {and}
  \bibinfo{person}{Jimmy Ba}.} \bibinfo{year}{2015}\natexlab{}.
\newblock \showarticletitle{Adam: A Method for Stochastic Optimization}. In
  \bibinfo{booktitle}{\emph{International Conference on Learning Representation
  (ICLR)}}.
\newblock


\bibitem[\protect\citeauthoryear{Kiros, Zhu, Salakhutdinov, Zemel, Torralba,
  Urtasun, and Fidler}{Kiros et~al\mbox{.}}{2015}]%
        {kiros2015}
\bibfield{author}{\bibinfo{person}{Ryan Kiros}, \bibinfo{person}{Yukun Zhu},
  \bibinfo{person}{Ruslan Salakhutdinov}, \bibinfo{person}{Richard~S. Zemel},
  \bibinfo{person}{Antonio Torralba}, \bibinfo{person}{Raquel Urtasun}, {and}
  \bibinfo{person}{Sanja Fidler}.} \bibinfo{year}{2015}\natexlab{}.
\newblock \showarticletitle{Skip-thought Vectors}. In
  \bibinfo{booktitle}{\emph{Advances in neural information processing systems}}
  (Montreal, Canada). \bibinfo{publisher}{MIT Press},
  \bibinfo{address}{Cambridge, MA, USA}, \bibinfo{pages}{3294--3302}.
\newblock


\bibitem[\protect\citeauthoryear{Lee, Chen, Hua, Hu, and He}{Lee
  et~al\mbox{.}}{2018}]%
        {lee18}
\bibfield{author}{\bibinfo{person}{Kuang-Huei Lee}, \bibinfo{person}{Xi Chen},
  \bibinfo{person}{Gang Hua}, \bibinfo{person}{Houdong Hu}, {and}
  \bibinfo{person}{Xiaodong He}.} \bibinfo{year}{2018}\natexlab{}.
\newblock \showarticletitle{Stacked Cross Attention for Image-Text Matching}.
  In \bibinfo{booktitle}{\emph{Proceedings of the European Conference on
  Computer Vision (ECCV)}}.
\newblock


\bibitem[\protect\citeauthoryear{Lin, Maire, Belongie, Bourdev, Girshick, Hays,
  Perona, Ramanan, Zitnick, and Dollár}{Lin et~al\mbox{.}}{2015}]%
        {coco}
\bibfield{author}{\bibinfo{person}{Tsung-Yi Lin}, \bibinfo{person}{Michael
  Maire}, \bibinfo{person}{Serge Belongie}, \bibinfo{person}{Lubomir Bourdev},
  \bibinfo{person}{Ross Girshick}, \bibinfo{person}{James Hays},
  \bibinfo{person}{Pietro Perona}, \bibinfo{person}{Deva Ramanan},
  \bibinfo{person}{C.~Lawrence Zitnick}, {and} \bibinfo{person}{Piotr
  Dollár}.} \bibinfo{year}{2015}\natexlab{}.
\newblock \bibinfo{title}{Microsoft COCO: Common Objects in Context}.
\newblock
\newblock
\showeprint[arxiv]{1405.0312}~[cs.CV]


\bibitem[\protect\citeauthoryear{Mar{\'{i}}n, Biswas, Ofli, Hynes, Salvador,
  Aytar, Weber, and Torralba}{Mar{\'{i}}n et~al\mbox{.}}{2019}]%
        {marin2019}
\bibfield{author}{\bibinfo{person}{Javier Mar{\'{i}}n}, \bibinfo{person}{Aritro
  Biswas}, \bibinfo{person}{Ferda Ofli}, \bibinfo{person}{Nicholas Hynes},
  \bibinfo{person}{Amaia Salvador}, \bibinfo{person}{Yusuf Aytar},
  \bibinfo{person}{Ingmar Weber}, {and} \bibinfo{person}{Antonio Torralba}.}
  \bibinfo{year}{2019}\natexlab{}.
\newblock \showarticletitle{{Recipe1M + : A Dataset for Learning Cross-Modal
  Embeddings for Cooking Recipes and Food Images}}.
\newblock \bibinfo{journal}{\emph{IEEE Transactions on Pattern Analysis and
  Machine Intelligence}} (\bibinfo{year}{2019}).
\newblock


\bibitem[\protect\citeauthoryear{Mikolov, Sutskever, Chen, Corrado, and
  Dean}{Mikolov et~al\mbox{.}}{2013}]%
        {mikolov2013}
\bibfield{author}{\bibinfo{person}{Tomas Mikolov}, \bibinfo{person}{Ilya
  Sutskever}, \bibinfo{person}{Kai Chen}, \bibinfo{person}{Greg~S Corrado},
  {and} \bibinfo{person}{Jeff Dean}.} \bibinfo{year}{2013}\natexlab{}.
\newblock \showarticletitle{Distributed representations of words and phrases
  and their compositionality}. In \bibinfo{booktitle}{\emph{Advances in neural
  information processing systems}}. \bibinfo{pages}{3111--3119}.
\newblock


\bibitem[\protect\citeauthoryear{Nishimura, Hashimoto, and Mori}{Nishimura
  et~al\mbox{.}}{2019}]%
        {Nishimura1}
\bibfield{author}{\bibinfo{person}{Taichi Nishimura}, \bibinfo{person}{Atsushi
  Hashimoto}, {and} \bibinfo{person}{Shinsuke Mori}.}
  \bibinfo{year}{2019}\natexlab{}.
\newblock \showarticletitle{Procedural Text Generation from a Photo Sequence}.
  In \bibinfo{booktitle}{\emph{International Conference on Natural Language
  Generation}}.
\newblock


\bibitem[\protect\citeauthoryear{Nishimura, Hashimoto, Ushiku, Kameko,
  Yamakata, and Mori}{Nishimura et~al\mbox{.}}{2021}]%
        {Nishimura2}
\bibfield{author}{\bibinfo{person}{Taichi Nishimura}, \bibinfo{person}{Atsushi
  Hashimoto}, \bibinfo{person}{Yoshitaka Ushiku}, \bibinfo{person}{Hirotaka
  Kameko}, \bibinfo{person}{Yoko Yamakata}, {and} \bibinfo{person}{Shinsuke
  Mori}.} \bibinfo{year}{2021}\natexlab{}.
\newblock \showarticletitle{Structure-Aware Procedural Text Generation From an
  Image Sequence}.
\newblock \bibinfo{journal}{\emph{IEEE Access}}  \bibinfo{volume}{9}
  (\bibinfo{year}{2021}), \bibinfo{pages}{2125--2141}.
\newblock
\urldef\tempurl%
\url{https://doi.org/10.1109/ACCESS.2020.3043452}
\showDOI{\tempurl}


\bibitem[\protect\citeauthoryear{Ott, Edunov, Baevski, Fan, Gross, Ng,
  Grangier, and Auli}{Ott et~al\mbox{.}}{2019}]%
        {ott2019fairseq}
\bibfield{author}{\bibinfo{person}{Myle Ott}, \bibinfo{person}{Sergey Edunov},
  \bibinfo{person}{Alexei Baevski}, \bibinfo{person}{Angela Fan},
  \bibinfo{person}{Sam Gross}, \bibinfo{person}{Nathan Ng},
  \bibinfo{person}{David Grangier}, {and} \bibinfo{person}{Michael Auli}.}
  \bibinfo{year}{2019}\natexlab{}.
\newblock \showarticletitle{fairseq: A Fast, Extensible Toolkit for Sequence
  Modeling}. In \bibinfo{booktitle}{\emph{Proceedings of NAACL-HLT 2019:
  Demonstrations}}.
\newblock


\bibitem[\protect\citeauthoryear{Pham, Guerrero, Li, and Pavlovic}{Pham
  et~al\mbox{.}}{2021}]%
        {chef2021}
\bibfield{author}{\bibinfo{person}{Hai~Xuan Pham}, \bibinfo{person}{Ricardo
  Guerrero}, \bibinfo{person}{Jiatong Li}, {and} \bibinfo{person}{Vladimir
  Pavlovic}.} \bibinfo{year}{2021}\natexlab{}.
\newblock \showarticletitle{{CHEF}: Cross-modal Hierarchical Embeddings for
  Food Domain Retrieval}. In \bibinfo{booktitle}{\emph{AAAI}}.
\newblock


\bibitem[\protect\citeauthoryear{Plummer, Wang, Cervantes, Caicedo,
  Hockenmaier, and Lazebnik}{Plummer et~al\mbox{.}}{2015}]%
        {flicker30k}
\bibfield{author}{\bibinfo{person}{Bryan~A. Plummer}, \bibinfo{person}{Liwei
  Wang}, \bibinfo{person}{Chris~M. Cervantes}, \bibinfo{person}{Juan~C.
  Caicedo}, \bibinfo{person}{Julia Hockenmaier}, {and}
  \bibinfo{person}{Svetlana Lazebnik}.} \bibinfo{year}{2015}\natexlab{}.
\newblock \showarticletitle{Flickr30k Entities: Collecting Region-to-Phrase
  Correspondences for Richer Image-to-Sentence Models}. In
  \bibinfo{booktitle}{\emph{Proceedings of the IEEE International Conference on
  Computer Vision (ICCV)}}.
\newblock


\bibitem[\protect\citeauthoryear{Reed, Akata, Yan, Logeswaran, Schiele, and
  Lee}{Reed et~al\mbox{.}}{2016}]%
        {reed16}
\bibfield{author}{\bibinfo{person}{Scott Reed}, \bibinfo{person}{Zeynep Akata},
  \bibinfo{person}{Xinchen Yan}, \bibinfo{person}{Lajanugen Logeswaran},
  \bibinfo{person}{Bernt Schiele}, {and} \bibinfo{person}{Honglak Lee}.}
  \bibinfo{year}{2016}\natexlab{}.
\newblock \showarticletitle{Generative Adversarial Text to Image Synthesis}. In
  \bibinfo{booktitle}{\emph{Proceedings of The 33rd International Conference on
  Machine Learning}} \emph{(\bibinfo{series}{Proceedings of Machine Learning
  Research}, Vol.~\bibinfo{volume}{48})},
  \bibfield{editor}{\bibinfo{person}{Maria~Florina Balcan} {and}
  \bibinfo{person}{Kilian~Q. Weinberger}} (Eds.). \bibinfo{publisher}{PMLR},
  \bibinfo{address}{New York, New York, USA}, \bibinfo{pages}{1060--1069}.
\newblock
\urldef\tempurl%
\url{http://proceedings.mlr.press/v48/reed16.html}
\showURL{%
\tempurl}


\bibitem[\protect\citeauthoryear{Salvador, Drozdzal, i~Nieto, and
  Romero}{Salvador et~al\mbox{.}}{2019}]%
        {salvador_2019}
\bibfield{author}{\bibinfo{person}{Amaia Salvador}, \bibinfo{person}{Michal
  Drozdzal}, \bibinfo{person}{Xavier~Giro i Nieto}, {and}
  \bibinfo{person}{Adriana Romero}.} \bibinfo{year}{2019}\natexlab{}.
\newblock \showarticletitle{Inverse Cooking: Recipe Generation from Food
  Images}. In \bibinfo{booktitle}{\emph{CVPR}}.
\newblock


\bibitem[\protect\citeauthoryear{Salvador, Gundogdu, Bazzani, and
  Donoser}{Salvador et~al\mbox{.}}{2021}]%
        {salvador2021}
\bibfield{author}{\bibinfo{person}{Amaia Salvador}, \bibinfo{person}{Erhan
  Gundogdu}, \bibinfo{person}{Loris Bazzani}, {and} \bibinfo{person}{Michael
  Donoser}.} \bibinfo{year}{2021}\natexlab{}.
\newblock \showarticletitle{Revamping Cross-Modal Recipe Retrieval with
  Hierarchical Transformers and Self-supervised Learning}. In
  \bibinfo{booktitle}{\emph{CVPR}}.
\newblock


\bibitem[\protect\citeauthoryear{Salvador, Hynes, Aytar, Marin, Ofli, Weber,
  and Torralba}{Salvador et~al\mbox{.}}{2017}]%
        {salvador2017}
\bibfield{author}{\bibinfo{person}{Amaia Salvador}, \bibinfo{person}{Nicholas
  Hynes}, \bibinfo{person}{Yusuf Aytar}, \bibinfo{person}{Javier Marin},
  \bibinfo{person}{Ferda Ofli}, \bibinfo{person}{Ingmar Weber}, {and}
  \bibinfo{person}{Antonio Torralba}.} \bibinfo{year}{2017}\natexlab{}.
\newblock \showarticletitle{Learning cross-modal embeddings for cooking recipes
  and food images}. In \bibinfo{booktitle}{\emph{Proceedings of the IEEE
  conference on computer vision and pattern recognition}}.
  \bibinfo{pages}{3020--3028}.
\newblock


\bibitem[\protect\citeauthoryear{Sennrich, Haddow, and Birch}{Sennrich
  et~al\mbox{.}}{2016}]%
        {sennrich2016}
\bibfield{author}{\bibinfo{person}{R. Sennrich}, \bibinfo{person}{B. Haddow},
  {and} \bibinfo{person}{A. Birch}.} \bibinfo{year}{2016}\natexlab{}.
\newblock \showarticletitle{Improving Neural Machine Translation Models with
  Monolingual Data}. In \bibinfo{booktitle}{\emph{ACL}}.
\newblock


\bibitem[\protect\citeauthoryear{Tai, Socher, and Manning}{Tai
  et~al\mbox{.}}{2015}]%
        {tai2015}
\bibfield{author}{\bibinfo{person}{Kai~Sheng Tai}, \bibinfo{person}{Richard
  Socher}, {and} \bibinfo{person}{Christopher~D. Manning}.}
  \bibinfo{year}{2015}\natexlab{}.
\newblock \showarticletitle{Improved Semantic Representations From
  Tree-Structured Long Short-Term Memory Networks}. In
  \bibinfo{booktitle}{\emph{Proceedings of the 53rd Annual Meeting of the
  Association for Computational Linguistics and the 7th International Joint
  Conference on Natural Language Processing (Volume 1: Long Papers)}}.
  \bibinfo{publisher}{Association for Computational Linguistics},
  \bibinfo{address}{Beijing, China}, \bibinfo{pages}{1556--1566}.
\newblock
\urldef\tempurl%
\url{https://doi.org/10.3115/v1/P15-1150}
\showDOI{\tempurl}


\bibitem[\protect\citeauthoryear{Teng, Lin, and Adamic}{Teng
  et~al\mbox{.}}{2012}]%
        {teng2012recipe}
\bibfield{author}{\bibinfo{person}{Chun-Yuen Teng}, \bibinfo{person}{Yu-Ru
  Lin}, {and} \bibinfo{person}{Lada~A Adamic}.}
  \bibinfo{year}{2012}\natexlab{}.
\newblock \showarticletitle{Recipe recommendation using ingredient networks}.
  In \bibinfo{booktitle}{\emph{Proceedings of the 4th Annual ACM Web Science
  Conference}}. ACM, \bibinfo{pages}{298--307}.
\newblock


\bibitem[\protect\citeauthoryear{Vaswani, Shazeer, Parmar, Uszkoreit, Jones,
  Gomez, Kaiser, and Polosukhin}{Vaswani et~al\mbox{.}}{2017}]%
        {Vaswani2017}
\bibfield{author}{\bibinfo{person}{Ashish Vaswani}, \bibinfo{person}{Noam
  Shazeer}, \bibinfo{person}{Niki Parmar}, \bibinfo{person}{Jakob Uszkoreit},
  \bibinfo{person}{Llion Jones}, \bibinfo{person}{Aidan~N. Gomez},
  \bibinfo{person}{Lukasz Kaiser}, {and} \bibinfo{person}{Illia Polosukhin}.}
  \bibinfo{year}{2017}\natexlab{}.
\newblock \showarticletitle{Attention Is All You Need}. In
  \bibinfo{booktitle}{\emph{Advances in neural information processing
  systems}}.
\newblock


\bibitem[\protect\citeauthoryear{Vinyals, Toshev, Bengio, and Erhan}{Vinyals
  et~al\mbox{.}}{2015}]%
        {Vinyals2015}
\bibfield{author}{\bibinfo{person}{Oriol Vinyals}, \bibinfo{person}{Alexander
  Toshev}, \bibinfo{person}{Samy Bengio}, {and} \bibinfo{person}{Dumitru
  Erhan}.} \bibinfo{year}{2015}\natexlab{}.
\newblock \showarticletitle{Show and tell: A neural image caption generator}.
  In \bibinfo{booktitle}{\emph{2015 IEEE Conference on Computer Vision and
  Pattern Recognition (CVPR)}}. \bibinfo{pages}{3156--3164}.
\newblock
\urldef\tempurl%
\url{https://doi.org/10.1109/CVPR.2015.7298935}
\showDOI{\tempurl}


\bibitem[\protect\citeauthoryear{Wang, Lin, Hoi, , and Miao}{Wang
  et~al\mbox{.}}{2020a}]%
        {wang_eccv20}
\bibfield{author}{\bibinfo{person}{Hao Wang}, \bibinfo{person}{Guosheng Lin},
  \bibinfo{person}{Steven C.~H. Hoi}, \bibinfo{person}{}, {and}
  \bibinfo{person}{Chunyan Miao}.} \bibinfo{year}{2020}\natexlab{a}.
\newblock \showarticletitle{Structure-Aware Generation Network for Recipe
  Generation from Images}. In \bibinfo{booktitle}{\emph{ECCV}}.
\newblock


\bibitem[\protect\citeauthoryear{Wang, Sahoo, Liu, Lim, and Hoi}{Wang
  et~al\mbox{.}}{2019}]%
        {wang2019}
\bibfield{author}{\bibinfo{person}{Hao Wang}, \bibinfo{person}{Doyen Sahoo},
  \bibinfo{person}{Chenghao Liu}, \bibinfo{person}{Ee-peng Lim}, {and}
  \bibinfo{person}{Steven~CH Hoi}.} \bibinfo{year}{2019}\natexlab{}.
\newblock \showarticletitle{Learning Cross-Modal Embeddings with Adversarial
  Networks for Cooking Recipes and Food Images}. In
  \bibinfo{booktitle}{\emph{Proceedings of the IEEE Conference on Computer
  Vision and Pattern Recognition}}. \bibinfo{pages}{11572--11581}.
\newblock


\bibitem[\protect\citeauthoryear{Wang, Sahoo, Liu, Shu, Achananuparp, peng Lim,
  and Hoi}{Wang et~al\mbox{.}}{2020b}]%
        {wang2020}
\bibfield{author}{\bibinfo{person}{Hao Wang}, \bibinfo{person}{Doyen Sahoo},
  \bibinfo{person}{Chenghao Liu}, \bibinfo{person}{Ke Shu},
  \bibinfo{person}{Palakorn Achananuparp}, \bibinfo{person}{Ee peng Lim}, {and}
  \bibinfo{person}{Steven C.~H. Hoi}.} \bibinfo{year}{2020}\natexlab{b}.
\newblock \bibinfo{title}{Cross-Modal Food Retrieval: Learning a Joint
  Embedding of Food Images and Recipes with Semantic Consistency and Attention
  Mechanism}.
\newblock
\newblock
\showeprint[arxiv]{2003.03955}~[cs.CV]


\bibitem[\protect\citeauthoryear{Wolf, Debut, Sanh, Chaumond, Delangue, Moi,
  Cistac, Rault, Louf, Funtowicz, and Brew}{Wolf et~al\mbox{.}}{2019}]%
        {huggingface}
\bibfield{author}{\bibinfo{person}{Thomas Wolf}, \bibinfo{person}{Lysandre
  Debut}, \bibinfo{person}{Victor Sanh}, \bibinfo{person}{Julien Chaumond},
  \bibinfo{person}{Clement Delangue}, \bibinfo{person}{Anthony Moi},
  \bibinfo{person}{Pierric Cistac}, \bibinfo{person}{Tim Rault},
  \bibinfo{person}{R{\'{e}}mi Louf}, \bibinfo{person}{Morgan Funtowicz}, {and}
  \bibinfo{person}{Jamie Brew}.} \bibinfo{year}{2019}\natexlab{}.
\newblock \showarticletitle{HuggingFace's Transformers: State-of-the-art
  Natural Language Processing}.
\newblock \bibinfo{journal}{\emph{CoRR}}  \bibinfo{volume}{abs/1910.03771}
  (\bibinfo{year}{2019}).
\newblock
\showeprint[arxiv]{1910.03771}
\urldef\tempurl%
\url{http://arxiv.org/abs/1910.03771}
\showURL{%
\tempurl}


\bibitem[\protect\citeauthoryear{Wu, Schuster, Chen, Le, Norouzi, Macherey,
  Krikun, Cao, Gao, Macherey, Klingner, Shah, Johnson, Liu, Kaiser, Gouws,
  Kato, Kudo, Kazawa, Stevens, Kurian, Patil, Wang, Young, Smith, Riesa,
  Rudnick, Vinyals, Corrado, Hughes, and Dean}{Wu et~al\mbox{.}}{2016}]%
        {Wu2016}
\bibfield{author}{\bibinfo{person}{Yonghui Wu}, \bibinfo{person}{Mike
  Schuster}, \bibinfo{person}{Zhifeng Chen}, \bibinfo{person}{Quoc~V. Le},
  \bibinfo{person}{Mohammad Norouzi}, \bibinfo{person}{Wolfgang Macherey},
  \bibinfo{person}{Maxim Krikun}, \bibinfo{person}{Yuan Cao},
  \bibinfo{person}{Qin Gao}, \bibinfo{person}{Klaus Macherey},
  \bibinfo{person}{Jeff Klingner}, \bibinfo{person}{Apurva Shah},
  \bibinfo{person}{Melvin Johnson}, \bibinfo{person}{Xiaobing Liu},
  \bibinfo{person}{Lukasz Kaiser}, \bibinfo{person}{Stephan Gouws},
  \bibinfo{person}{Yoshikiyo Kato}, \bibinfo{person}{Taku Kudo},
  \bibinfo{person}{Hideto Kazawa}, \bibinfo{person}{Keith Stevens},
  \bibinfo{person}{George Kurian}, \bibinfo{person}{Nishant Patil},
  \bibinfo{person}{Wei Wang}, \bibinfo{person}{Cliff Young},
  \bibinfo{person}{Jason Smith}, \bibinfo{person}{Jason Riesa},
  \bibinfo{person}{Alex Rudnick}, \bibinfo{person}{Oriol Vinyals},
  \bibinfo{person}{Greg Corrado}, \bibinfo{person}{Macduff Hughes}, {and}
  \bibinfo{person}{Jeffrey Dean}.} \bibinfo{year}{2016}\natexlab{}.
\newblock \showarticletitle{Google's Neural Machine Translation System:
  Bridging the Gap between Human and Machine Translation}.
\newblock \bibinfo{journal}{\emph{CoRR}}  \bibinfo{volume}{abs/1609.08144}
  (\bibinfo{year}{2016}).
\newblock
\showeprint[arxiv]{1609.08144}
\urldef\tempurl%
\url{http://arxiv.org/abs/1609.08144}
\showURL{%
\tempurl}


\bibitem[\protect\citeauthoryear{Xie, Dai, Hovy, Luong, and Le}{Xie
  et~al\mbox{.}}{2020}]%
        {xie2020}
\bibfield{author}{\bibinfo{person}{Q. Xie}, \bibinfo{person}{Z. Dai},
  \bibinfo{person}{E. Hovy}, \bibinfo{person}{M. Luong}, {and}
  \bibinfo{person}{Q. Le}.} \bibinfo{year}{2020}\natexlab{}.
\newblock \showarticletitle{Unsupervised Data Augmentation for Consistency
  Training}. In \bibinfo{booktitle}{\emph{NeurIPS}}.
\newblock


\bibitem[\protect\citeauthoryear{Xu, Zhang, Huang, Zhang, Gan, Huang, and
  He}{Xu et~al\mbox{.}}{2018}]%
        {Xu_2018_CVPR}
\bibfield{author}{\bibinfo{person}{Tao Xu}, \bibinfo{person}{Pengchuan Zhang},
  \bibinfo{person}{Qiuyuan Huang}, \bibinfo{person}{Han Zhang},
  \bibinfo{person}{Zhe Gan}, \bibinfo{person}{Xiaolei Huang}, {and}
  \bibinfo{person}{Xiaodong He}.} \bibinfo{year}{2018}\natexlab{}.
\newblock \showarticletitle{AttnGAN: Fine-Grained Text to Image Generation With
  Attentional Generative Adversarial Networks}. In
  \bibinfo{booktitle}{\emph{Proceedings of the IEEE Conference on Computer
  Vision and Pattern Recognition (CVPR)}}.
\newblock


\bibitem[\protect\citeauthoryear{Zhang, Xu, Li, Zhang, Wang, Huang, and
  Metaxas}{Zhang et~al\mbox{.}}{2017a}]%
        {han2017stackgan}
\bibfield{author}{\bibinfo{person}{Han Zhang}, \bibinfo{person}{Tao Xu},
  \bibinfo{person}{Hongsheng Li}, \bibinfo{person}{Shaoting Zhang},
  \bibinfo{person}{Xiaogang Wang}, \bibinfo{person}{Xiaolei Huang}, {and}
  \bibinfo{person}{Dimitris Metaxas}.} \bibinfo{year}{2017}\natexlab{a}.
\newblock \showarticletitle{StackGAN: Text to Photo-realistic Image Synthesis
  with Stacked Generative Adversarial Networks}. In
  \bibinfo{booktitle}{\emph{Proceedings of the IEEE/CVF International
  Conference on Computer Vision (ICCV)}}.
\newblock


\bibitem[\protect\citeauthoryear{Zhang, Xu, Li, Zhang, Wang, Huang, and
  Metaxas}{Zhang et~al\mbox{.}}{2017b}]%
        {Zhang2017-jm}
\bibfield{author}{\bibinfo{person}{Han Zhang}, \bibinfo{person}{Tao Xu},
  \bibinfo{person}{Hongsheng Li}, \bibinfo{person}{Shaoting Zhang},
  \bibinfo{person}{Xiaogang Wang}, \bibinfo{person}{Xiaolei Huang}, {and}
  \bibinfo{person}{Dimitris~N Metaxas}.} \bibinfo{year}{2017}\natexlab{b}.
\newblock \showarticletitle{Stackgan: Text to photo-realistic image synthesis
  with stacked generative adversarial networks}. In
  \bibinfo{booktitle}{\emph{Proceedings of the {IEEE} international conference
  on computer vision}}. \bibinfo{pages}{5907--5915}.
\newblock


\bibitem[\protect\citeauthoryear{Zhu and Ngo}{Zhu and Ngo}{2020}]%
        {zhu2020}
\bibfield{author}{\bibinfo{person}{B. Zhu} {and} \bibinfo{person}{Chong-Wah
  Ngo}.} \bibinfo{year}{2020}\natexlab{}.
\newblock \showarticletitle{{CookGAN}: Causality based Text-to-Image
  Synthesis}. In \bibinfo{booktitle}{\emph{CVPR}}.
\newblock


\bibitem[\protect\citeauthoryear{Zhu, Ngo, Chen, and Hao}{Zhu
  et~al\mbox{.}}{2019a}]%
        {Zhu2019}
\bibfield{author}{\bibinfo{person}{Bin Zhu}, \bibinfo{person}{Chong-Wah Ngo},
  \bibinfo{person}{Jingjing Chen}, {and} \bibinfo{person}{Yanbin Hao}.}
  \bibinfo{year}{2019}\natexlab{a}.
\newblock \showarticletitle{R2GAN: Cross-Modal Recipe Retrieval With Generative
  Adversarial Network}. In \bibinfo{booktitle}{\emph{Proceedings of the
  IEEE/CVF Conference on Computer Vision and Pattern Recognition (CVPR)}}.
\newblock


\bibitem[\protect\citeauthoryear{Zhu, Pan, Chen, and Yang}{Zhu
  et~al\mbox{.}}{2019b}]%
        {Zhu_2019_CVPR}
\bibfield{author}{\bibinfo{person}{Minfeng Zhu}, \bibinfo{person}{Pingbo Pan},
  \bibinfo{person}{Wei Chen}, {and} \bibinfo{person}{Yi Yang}.}
  \bibinfo{year}{2019}\natexlab{b}.
\newblock \showarticletitle{DM-GAN: Dynamic Memory Generative Adversarial
  Networks for Text-To-Image Synthesis}. In
  \bibinfo{booktitle}{\emph{Proceedings of the IEEE/CVF Conference on Computer
  Vision and Pattern Recognition (CVPR)}}.
\newblock


\bibitem[\protect\citeauthoryear{Zhu, Sobihani, and Guo}{Zhu
  et~al\mbox{.}}{2015}]%
        {zhu2015}
\bibfield{author}{\bibinfo{person}{Xiaodan Zhu}, \bibinfo{person}{Parinaz
  Sobihani}, {and} \bibinfo{person}{Hongyu Guo}.}
  \bibinfo{year}{2015}\natexlab{}.
\newblock \showarticletitle{{Long Short-Term Memory Over Recursive
  Structures}}. In \bibinfo{booktitle}{\emph{Proceedings of the 32nd
  International Conference on Machine Learning}}
  \emph{(\bibinfo{series}{Proceedings of Machine Learning Research},
  Vol.~\bibinfo{volume}{37})}, \bibfield{editor}{\bibinfo{person}{Francis Bach}
  {and} \bibinfo{person}{David Blei}} (Eds.). \bibinfo{publisher}{PMLR},
  \bibinfo{address}{Lille, France}, \bibinfo{pages}{1604--1612}.
\newblock
\urldef\tempurl%
\url{http://proceedings.mlr.press/v37/zhub15.html}
\showURL{%
\tempurl}


\end{thebibliography}

\onecolumn
\section*{\centering{Supplementary Material}}
\vspace{3em}

\begin{multicols}{2}

\subsection*{Cross-modal Retrieval}

\begin{figure*}
\centering
\includegraphics[width=0.95\textwidth]{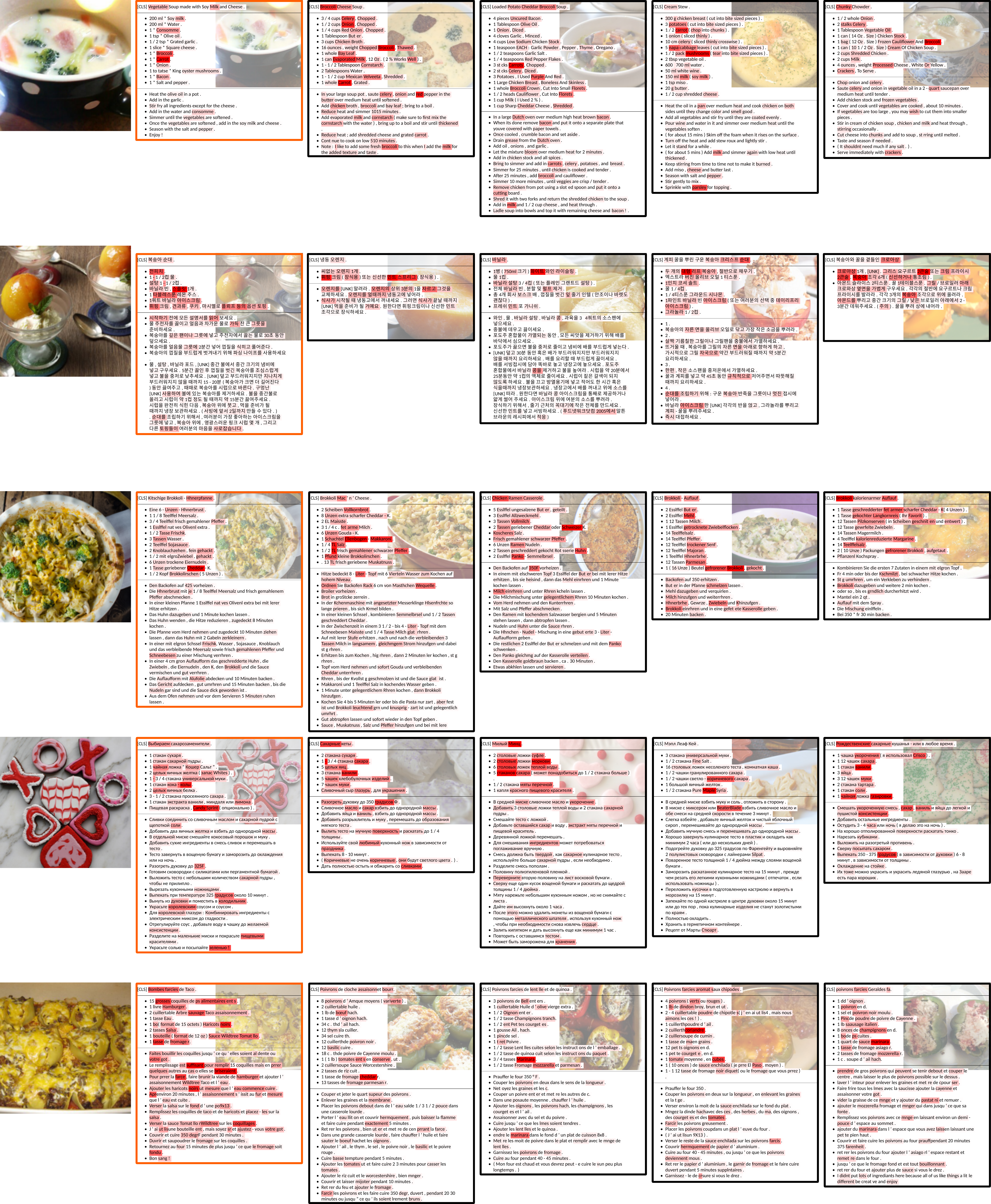}
\vspace{-1em}
\caption{Image-to-recipe ``multilingual'' retrieval. Top five retrieved recipes are shown for each query image. Highlighted recipes (leftmost) are the exact (top-1) matches of corresponding query images.
Notice that recipe attention primarily focuses on words that help describe the visual appearance of the query.}
\label{fig:i2r}
\end{figure*}

\begin{figure*}
\centering
\includegraphics[width=\textwidth]{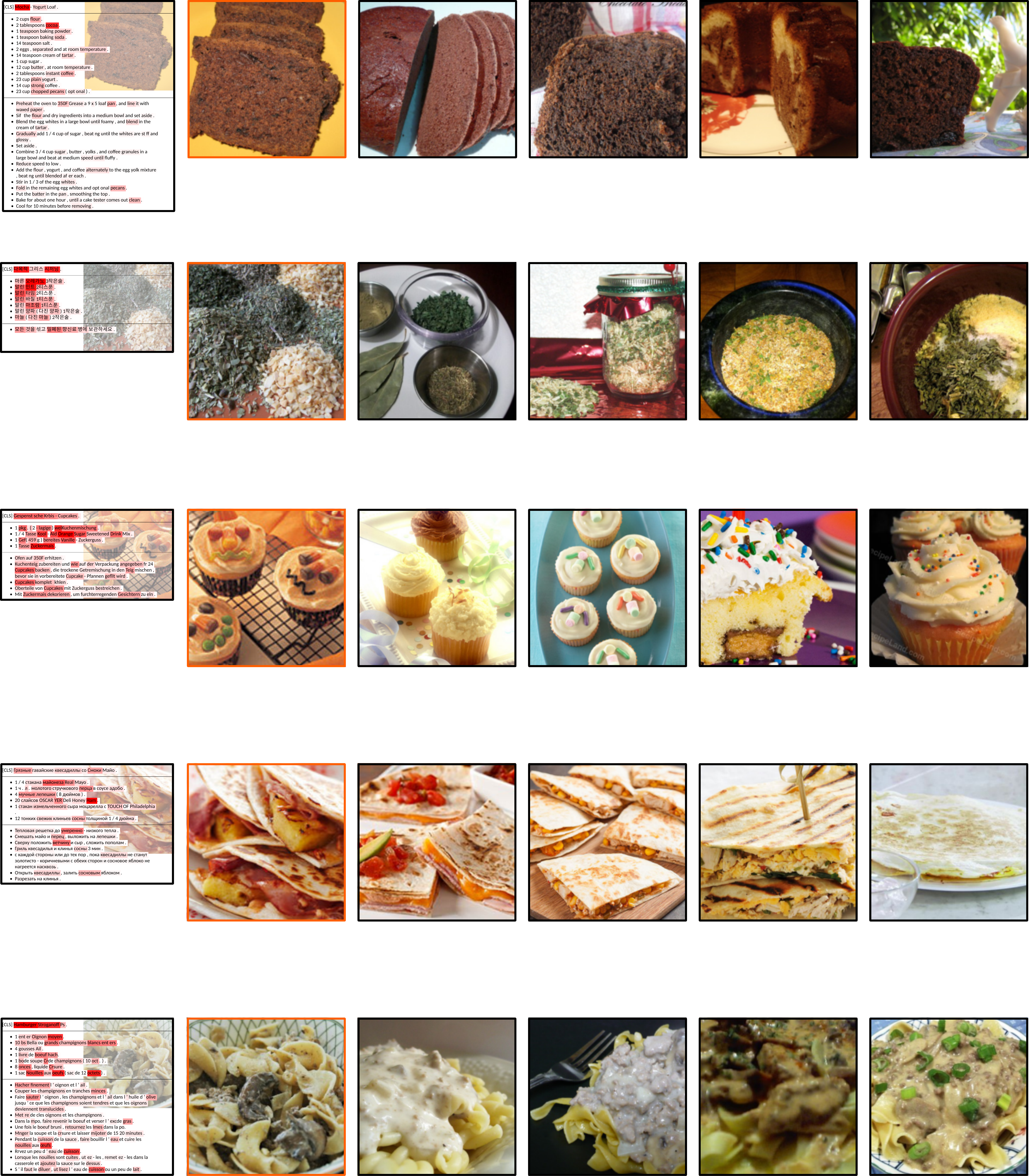}
\caption{Recipe-to-image ``multilingual'' retrieval. Top five retrieved images are shown for each query recipe. Highlighted images (leftmost) are the exact (top-1) matches of corresponding query recipes.
Notice that recipe attention primarily focuses on words that help describe the visual appearance of the query.}
\label{fig:r2i}
\end{figure*}

\autoref{fig:i2r} and \autoref{fig:r2i} show ``multilingual'' retrieval results from our T-ML model. In these figures, each row corresponds to a query (image in \autoref{fig:i2r} and recipe in \autoref{fig:r2i}), followed by the top five cross-modal retrieved samples in each of the five supported languages (English, Korean, German, Russian and French). Note that since our recipe embedding network is based on transformer~\cite{Vaswani2017} encoders, we can implicitly obtain attention over the different words in a recipe. Note that: 
\begin{itemize}
    \item Word-piece attention calculated as proposed in \cite{Abnar2020}.
    \item Words reconstructed from corresponding word-pieces.
    \item Word attention obtained as sum over word-pieces.
    \item Displayed attention reflects that of the `[CLS]' token.
    \item `[CLS]' attention omitted.
\end{itemize}

Additionally, in \autoref{fig:i2r} and \autoref{fig:r2i} it can be appreciated that the recipe encoder focuses primarily on the ingredients section, and in particular mostly on those words that help describe the visual appearance of the query.

\subsubsection*{\textbf{Incomplete recipe analysis.}}
\begin{figure*}
\centering
\includegraphics[width=\textwidth]{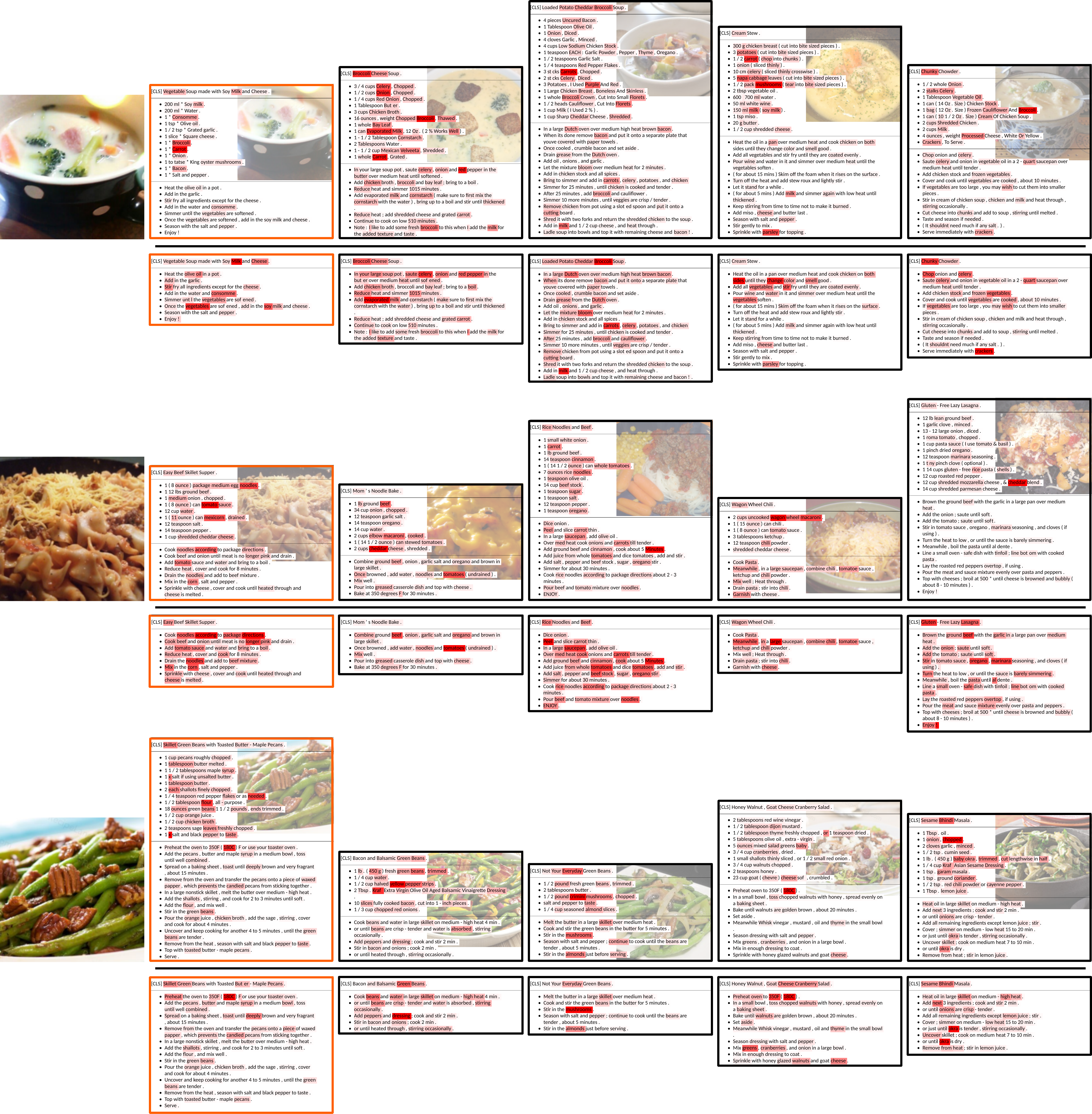}
\caption{Image-to-recipe retrieval. Recipe attention of our T-ML model with and without the use of ingredient information. Top-5 retrieved recipes shown for each query image. Highlighted recipes (leftmost) are the exact (top-1) matches of corresponding query images.
Notice that for both complete and incomplete recipes attention primarily focuses on words that help describe the visual appearance of the query.}
\label{fig:i2r_title-inst}
\end{figure*}

In this experiment we test our T-ML retrieval model (trained on the full recipe information) on incomplete/partial data. We explore the amount of information contributed by title, ingredient lines and instructions toward retrieval performance. \autoref{tab:ablation} indicates that, ingredients appear to be the most important piece of information when tested on a model trained on complete data. Interestingly, this differs from models trained with incomplete recipe information, where it was found that instructions carried the most importance. This observation, along with the poor performance of the model trained only on title, leads to hypothesize that those models are not effectively trained. \autoref{fig:i2r_title-inst} shows three image-to-recipe retrieval examples. There, it can be appreciated that in the case of complete recipe information, the recipe encoder again focuses primarily on the ingredients section, and in particular mostly on those words that help describe the visual appearance of the query, e.g. see top left, where the full recipe focuses on `broccoli', `carrot' and `milk'. On the other hand, when recipes are missing the ingredients information, attention is redistributed to other words that also help describe the visual appearance of the query, e.g. see top left, where the incomplete recipe focuses on `vegetables', `milk' and `soy' (notice the words `broccoli' and `carrot' no longer appear, thus extra emphasis is placed on `vegetables').

\begin{table}[!ht]
  \centering
  \small
  \caption{Ablation analysis of textual recipe components. Performance metrics reported for 1k search space. ``Tr'' indicates whether the model was trained using only partial inputs. }
  \begin{tabular}{l|l|cccc}
\toprule
	 \multicolumn{6}{c}{\textbf{Image-to-Recipe Retrieval}} \\ \hline
	\textbf{Input} & \textbf{Tr}  & \textbf{medR} & \textbf{R1} & \textbf{R5} & \textbf{R10} \\ 
	\midrule
	\multirow{2}*{title} & \multirow{1}*{Yes}  & 29.7 & 8.4 & 22.8 & 32.1 \\ \cmidrule{2-6}
	                     & \multirow{1}*{No}   & 8.9 & 19.5 & 42.52 & 52.74 \\ \cmidrule{1-6}
	                     
	\multirow{2}*{ingr} & \multirow{1}*{Yes}  & 3 & 36.0 & 63.2 & 73.4 \\ \cmidrule{2-6}
	                     & \multirow{1}*{No}  & 2 & 43.15 & 70.07 & 78.51 \\  \cmidrule{1-6}
	                     
	\multirow{2}*{inst} & \multirow{1}*{Yes}  & 2 & 48.1 & 76.7 & 84.8 \\ \cmidrule{2-6}
	                     & \multirow{1}*{No}  & 2 & 40.25 & 67.91 & 76.75 \\ \cmidrule{1-6}
	                     
	\multirow{2}*{\begin{tabular}{@{}c@{}}title, \\ ingr\end{tabular}} & \multirow{1}*{Yes} & 2 & 45.4 & 74.4 & 82.6 \\ \cmidrule{2-6}
	                                                                   & \multirow{1}*{No}  & 1 & 53.93 & 80.47 & 87.27 \\ \cmidrule{1-6}
	                     
	\multirow{2}*{\begin{tabular}{@{}c@{}}title, \\ inst\end{tabular}} & \multirow{1}*{Yes} & 1 & 51.6 & 79.8 & 87.1 \\  \cmidrule{2-6}
	                                                                   & \multirow{1}*{No}  & 1.9 & 48.56 & 77.15 & 84.77 \\  \cmidrule{1-6}
	                     
	\multirow{2}*{\begin{tabular}{@{}c@{}}ingr, \\ inst\end{tabular}} & \multirow{1}*{Yes} & 1 & 62.2 & 86.6 & 91.6 \\ \cmidrule{2-6}
	                                                                  & \multirow{1}*{No}  & 1 & 61.89 & 85.94 & 91.14 \\  
	\bottomrule
     \end{tabular}
  \label{tab:ablation}
 \end{table}

\subsection*{Cross-modal Synthesis}

In this section we give additional results of using our retrieval framework along with image generation.

\subsubsection*{\textbf{Recipe-to-image synthesis}.}
\autoref{fig:r2i_i2i_syn} depicts the synthetics generated from embeddings of ACME~\cite{wang2019}, CHEF~\cite{chef2021} and our transformer-based models. It is the expansion of Fig. 2 in the main paper, where we also include images generated from the paired visual embeddings of each respective example. It can be observed that the images generated from both modaltity embeddings of our proposed T and T-ML models have similar visual attributes (Recall that all models have been trained on the recipe-to-image task only). In other words, the two modality embeddings of our proposed model are more closely aligned in the representation.

\begin{figure*}[!ht]
\centering
\includegraphics[width=\textwidth]{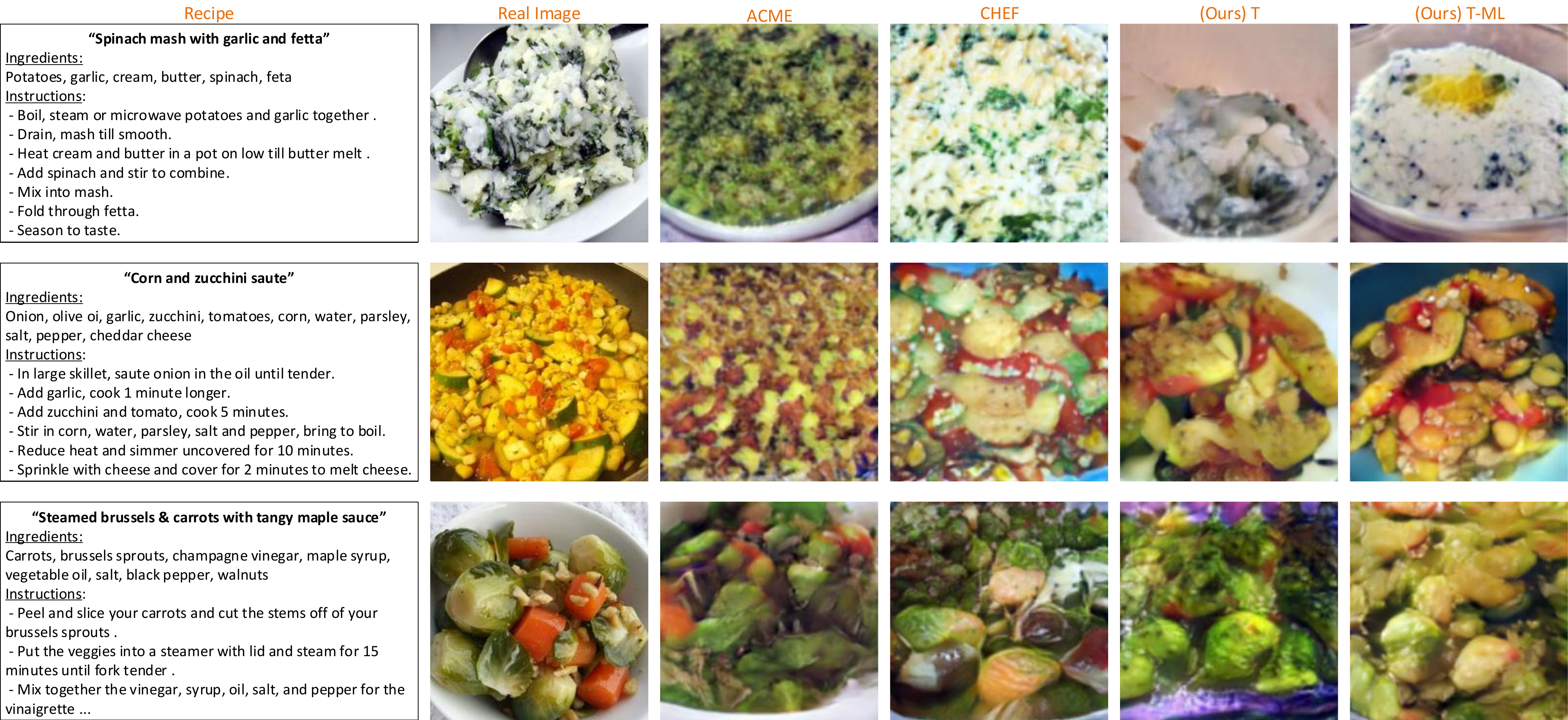}
\caption{Recipe-to-image synthesis. In this figure, real and synthetic images can successfully retrieve the real recipes.}
\label{fig:good_real_good_syn}
\end{figure*}

\begin{figure*}[!ht]
\centering
\includegraphics[width=\textwidth]{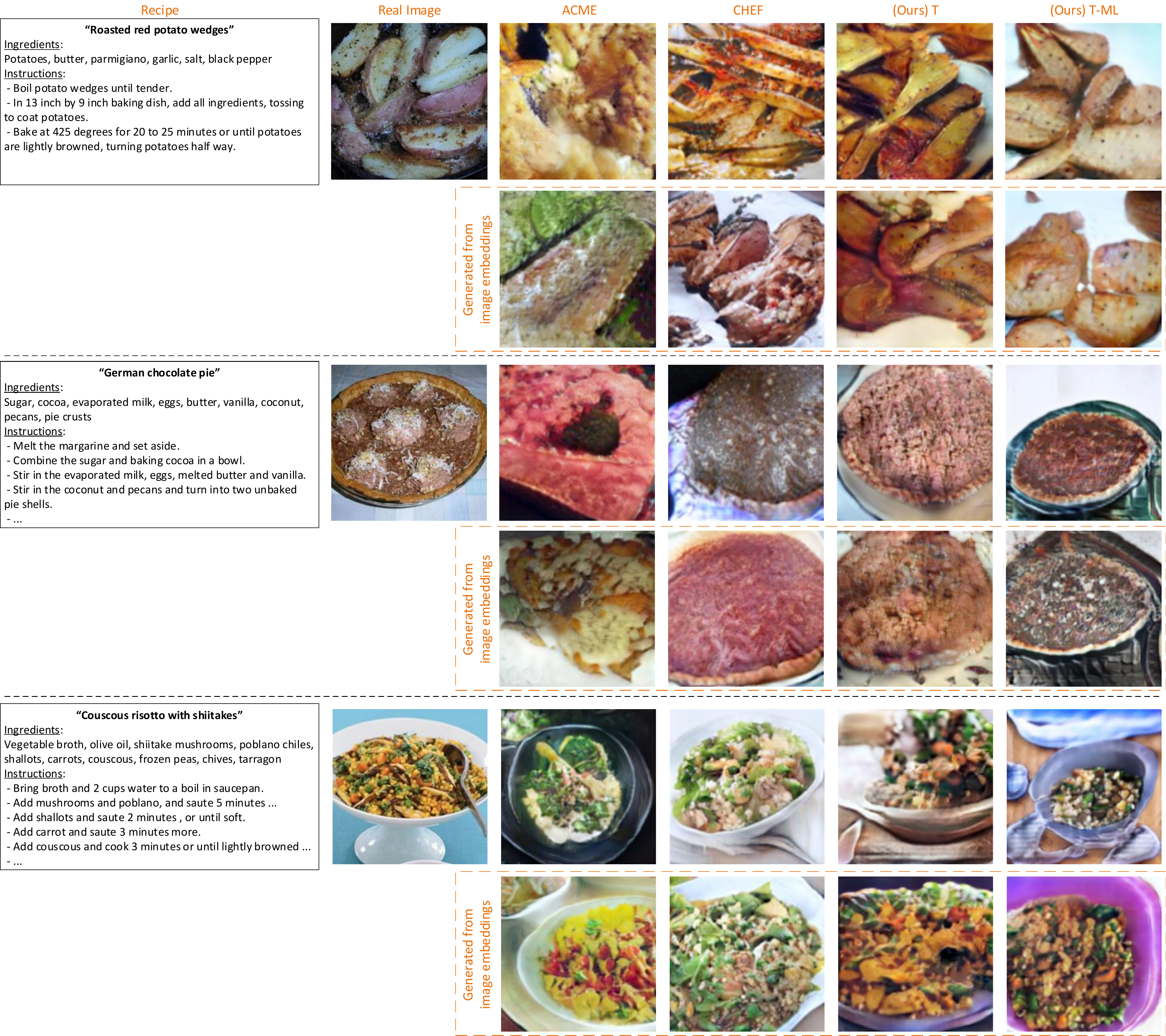}
\caption{Image synthesis from recipe and image embeddings. For each input recipe/image, the first row shows synthetic images created from recipe embedding, while the second row shows images generated from image embedding. It is the expansion of Fig. \ref{fig:synthesis} in the main paper. Notice the images generated from both modality embeddings of our proposed model have similar visual attributes.}
\label{fig:r2i_i2i_syn}
\end{figure*}

\begin{figure*}[!ht]
\centering
\includegraphics[width=\textwidth]{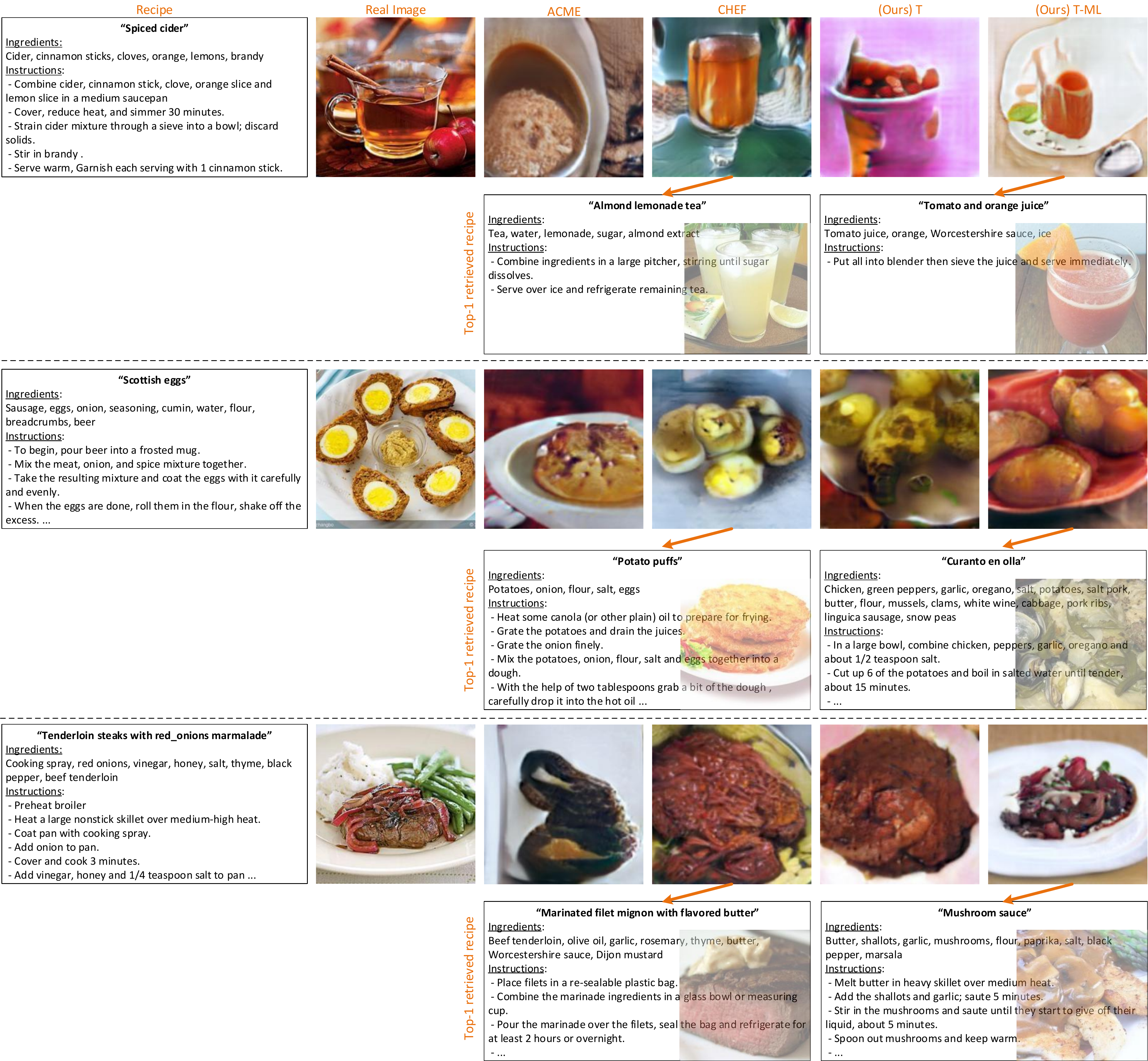}
\caption{Recipe-to-image synthesis, where synthetic images fail to match the real recipes, however the corresponding real images can retrieve the recipes successfully. For each example, the second row shows the top-1 recipes retrieved using the images generated from CHEF and T-ML embeddings.
}
\label{fig:good_real_bad_syn}
\end{figure*}

\begin{figure*}[!ht]
\centering
\includegraphics[width=\textwidth]{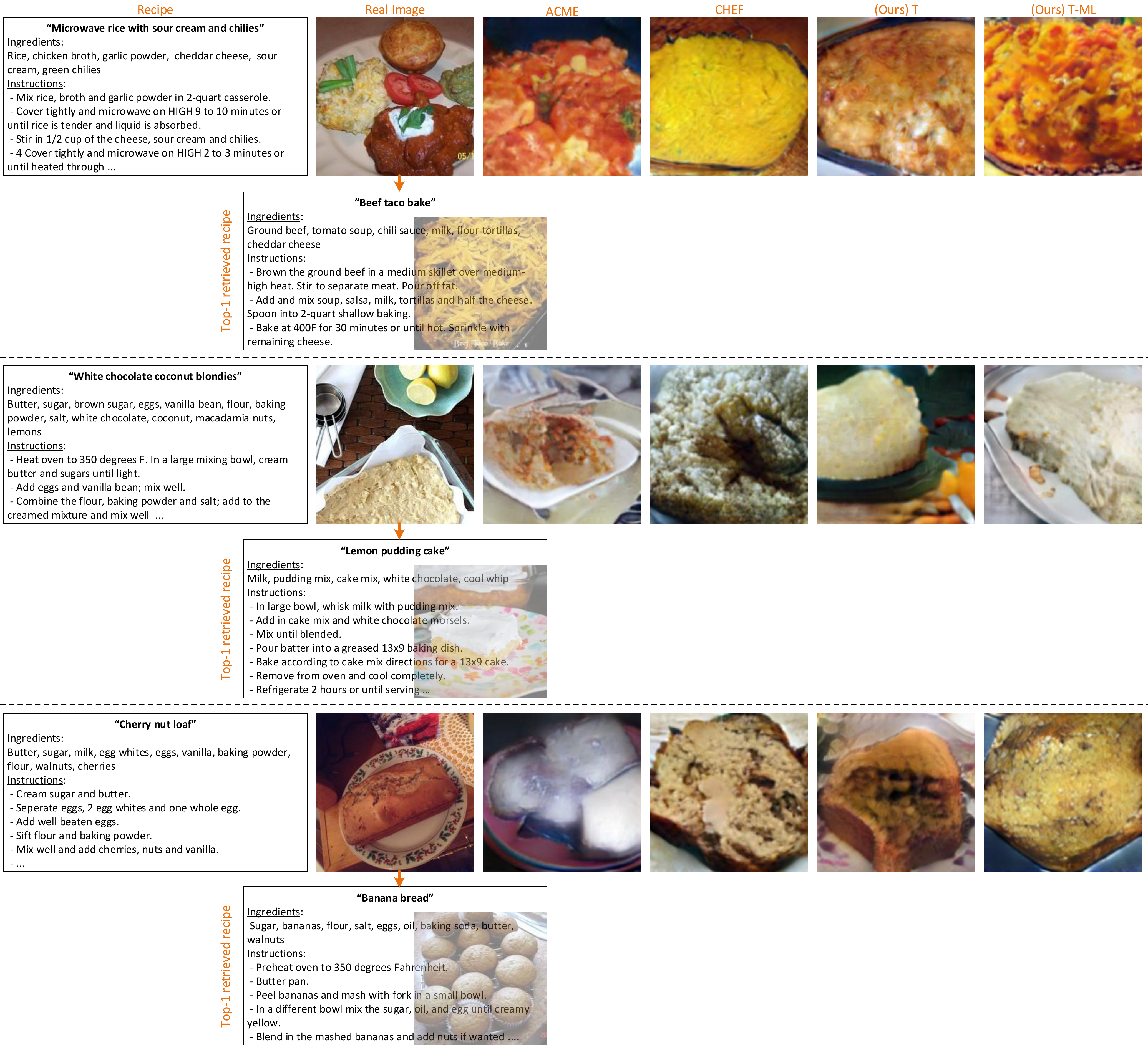}
\caption{Recipe-to-image synthesis, where synthetic images can be used to successfully retrieve the real recipes, however the corresponding real images fail to retrieve the recipes. For each example, the second row shows the top-1 recipe retrieved using the real image. 
}
\label{fig:bad_real_good_syn}
\end{figure*}

\autoref{fig:good_real_good_syn} depicts some examples of recipe-to-image synthesis where both real and synthetic images manage to retrieve the correct recipes. \autoref{fig:good_real_bad_syn} shows other recipe-to-image synthetically generated examples where synthetic images fail to retrieve the correct recipes, while real images are successful. \autoref{fig:bad_real_good_syn} illustrates examples where real images fail to retrieve the correct recipe pair, while synthetic images (recipe-to-image) are successful. Particularly, in the first and second examples, the main dishes are not of the main focus in the original images, while in the last example it is very difficult to identity the type of bread loaf. However, the generative models (from CHEF, T-ML) manage to synthesize images that perfectly match the real recipes. These results suggest the potential of using GAN to regularize the ``difficult'' paired examples while training the retrieval model, which may bring about performance gain.

\subsubsection*{\textbf{Recipe-to-image synthesis variation.}}
\begin{figure*}[!ht]
\centering
\includegraphics[width=\textwidth]{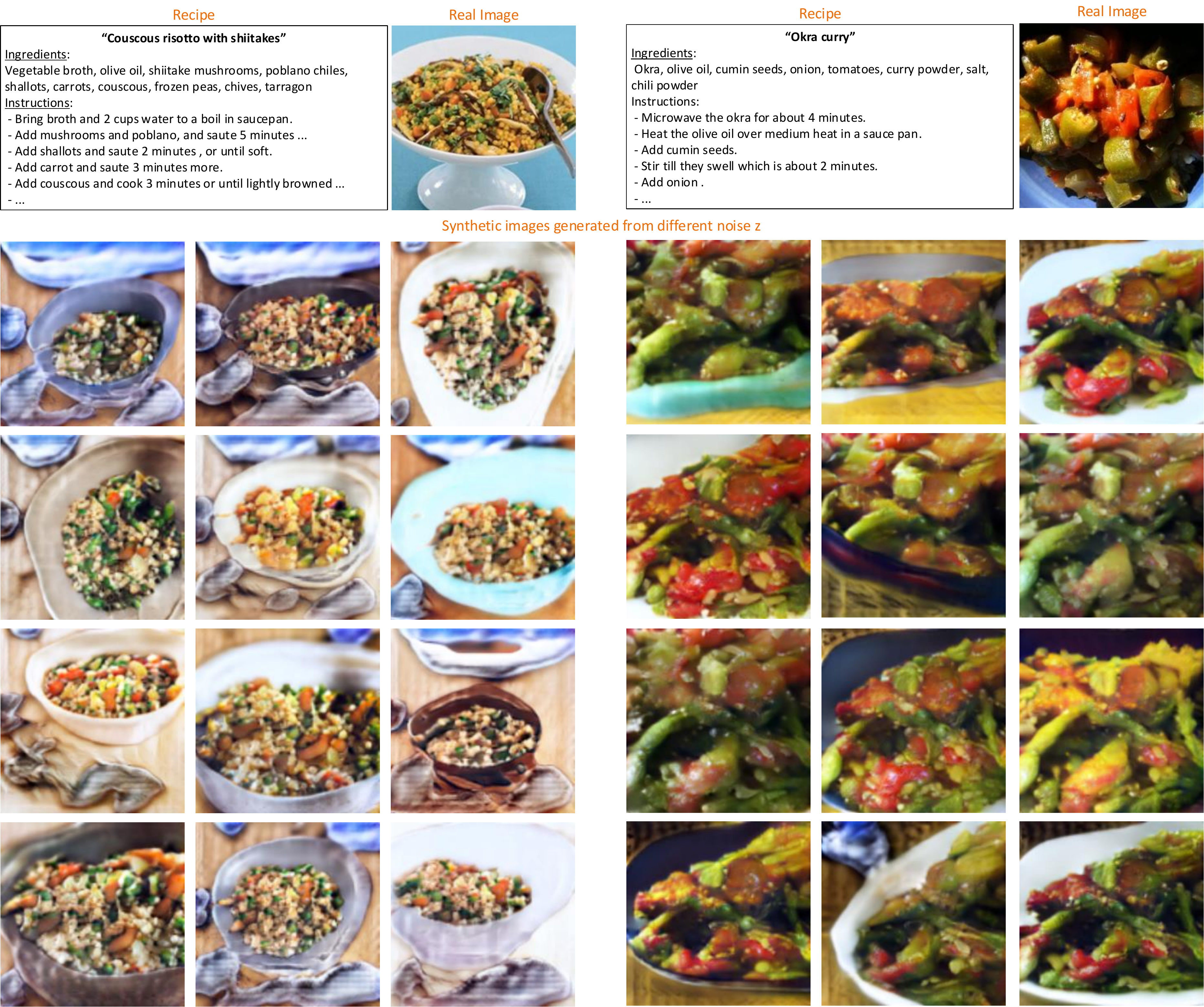}
\caption{Image synthesis from recipe with different noise input $z$, using our T-ML embedding and generator. The top row includes real recipes and their corresponding original images. The subsequent rows show synthetic images created from the same recipes, with different noise input vectors $z$. The attributes of the dish (color, ingredient pieces) are consistent, but the layout varies.}
\label{fig:syn_multi_noise}
\end{figure*}

\autoref{fig:syn_multi_noise} shows images generated from two different recipes, as well as different input noise $z$, using our proposed T-ML embedding and generator. The main attributes of the dish (color, ingredient pieces) are consistent, but the viewpoint, position and background vary. Thus, the input noise can be considered as ``style'' factor of the synthetic image,  different style factors lead to the varying visual appearances of images generated from the same recipe.



\end{multicols}

\end{document}